\documentclass[10pt,journal,compsoc]{IEEEtran}
\IEEEoverridecommandlockouts
\usepackage{amsmath,amsfonts}
\usepackage{algorithmic}
\usepackage{algorithm}
\usepackage{array}
\usepackage[caption=false,font=normalsize,labelfont=sf,textfont=sf]{subfig}
\usepackage{textcomp}
\usepackage{stfloats}
\usepackage{url}
\usepackage{verbatim}
\usepackage{graphicx}
\usepackage{cite}
\usepackage{multirow}
\usepackage{multicol}
\usepackage{booktabs}
\usepackage{bm}
\usepackage{wrapfig}
\usepackage[table,dvipsnames]{xcolor}
\usepackage{pifont}
\usepackage{makecell}
\usepackage{enumitem}
\usepackage{balance}
\usepackage{ragged2e}
\usepackage{float,color}
\usepackage{threeparttable}
\usepackage{colortbl}
\usepackage{caption}
\usepackage[percent]{overpic}
\usepackage{helvet}

\usepackage{tikz}
\usepackage[edges]{forest}
\definecolor{rootcolor}{RGB}{0, 104, 139}
\definecolor{level1color}{RGB}{72, 145, 173}
\definecolor{level2color}{RGB}{144, 186, 201}
\definecolor{level3color}{RGB}{216, 227, 231}
\definecolor{textcolor}{RGB}{51, 51, 51}
\definecolor{mydarkblue}{rgb}{0,0.08,0.45}

\def\etal{\emph{et al}.}
\def\eg{\emph{e.g}.} 
\def\ie{\emph{i.e}.} 
 
 \def\vs{\emph{vs}.}

\newcommand{\cmark}{\ding{51}}%
\newcommand{\xmark}{\ding{55}}%

\newcommand{\paratitle}[1]{\vspace{0.5ex}\noindent\textbf{\textit{#1}}}

\usepackage[colorlinks, linkcolor=red, citecolor=blue, urlcolor=blue]{hyperref}

\begin{document}
\bstctlcite{IEEEtranBSTCTL:dash_repeated_names}

\title{From Question Answering to Task Completion: A Survey on Agent System and Harness Design}

\author{Jianyuan Guo, Zhiwei Hao, Chengcheng Wang, Cheng Fan, Tingzhang Luo, Hongguang Li, Ying Gao, Hefei Mei, Jiankun Peng, Rongjian Xu, Minjing Dong, Han Wu, Mengyu Zheng, Kai Han, Shiqi~Wang, Chang~Xu and Yunhe Wang
\IEEEcompsocitemizethanks{
\IEEEcompsocthanksitem J. Guo, Z. Hao, C. Fan, T. Luo, H. Li, Y. Gao, H. Mei, J. Peng, R. Xu, M. Dong and S. Wang are with the Department of Computer Science, City University
of Hong Kong, HKSAR, China. Email: \{jianyguo, zhiwei.hao, minjdong, shiqwang\}@cityu.edu.hk.
\IEEEcompsocthanksitem C. Wang and C. Xu are with the School of Computer Science, University of Sydney. Email: cwan0785@uni.sydney.edu.au, c.xu@sydney.edu.au.
\IEEEcompsocthanksitem H. Wu is with the Peking University. Email: han.wu@pku.edu.cn.
\IEEEcompsocthanksitem M. Zheng, K. Han and Y. Wang are with the TokenRhythm Technologies. Email: \{mengyu.zheng, kai.han, yunhe.wang\}@tokenrhythm.ai.
\IEEEcompsocthanksitem Correspondence author: Chang Xu and Yunhe Wang.
}}

\IEEEtitleabstractindextext{

\begin{abstract}\justifying
LLM-based agents mark a shift from passive question answering to active task completion: they perceive environments, invoke tools, maintain state, and act over extended horizons.
As agent systems have evolved from prompt engineering to workflows and context engineering, harness engineering, and agent-native training with co-evolution, a central question has become increasingly important: where does the bottleneck in agent performance reside—in the foundation model, in the execution harness, or in the coupling between them?
This survey examines LLM-based agents through a model--harness lens.
We first clarify the functional definition of agents and the implementation view of an LLM-based agent as a foundation model coupled with an execution harness.
We then analyze the limits of model-centric scaling, trace four paradigms of agent engineering, and decompose the execution harness into six coupled runtime responsibilities: observation, context, control, action, state, and verification/governance.
Using this decomposition, we map task properties and domain pressures to harness configurations, review benchmark and evaluation practices, and synthesize model--harness evidence on how runtime design affects long-horizon task completion, efficiency, and reliability.
Finally, we identify open challenges in value-aware evaluation, safety, harness generalization, and model--harness co-evolution.
Rather than treating agents as models with auxiliary tools, this survey argues that agent quality---including success, efficiency, safety, and generalization---emerges from the interaction between model capability, runtime infrastructure, task structure, and evaluation design.
A collection of papers discussed in this survey is provided in \href{https://github.com/ggjy/Awesome-Agent-Engineering}{https://github.com/ggjy/Awesome-Agent-Engineering}.
\end{abstract}

}

\maketitle

\IEEEdisplaynontitleabstractindextext

\section{Introduction}
\label{sec:intro}

\begin{quote}
\textit{``Nothing is particularly hard if you divide it into small jobs.''}
\hfill --- Henry Ford
\end{quote}

\IEEEPARstart{L}{arge} language model (LLM)-based agents---autonomous systems that perceive environments, reason over goals, and execute multi-step actions---mark a transition from passive question answering~\cite{brown2020gpt3,ouyang2022instructgpt} to active task completion~\cite{luo2025llmagent,xi2025rise}.
Unlike early chat interfaces that optimized single-turn response quality, modern agent systems operate as closed loops that invoke tools, update state, and verify outcomes over extended horizons.
Prominent examples span multiple domains: coding agents such as Devin~\cite{cognition2024devin}, Claude Code~\cite{claudecode2025}, and Codex~\cite{openai2026harness} independently diagnose and resolve software engineering tasks across entire repositories; general-purpose agents like Manus~\cite{shen2025manus} orchestrate multi-step workflows from research to data analysis; open-source platforms including AutoGPT~\cite{richards2023autogpt}, OpenHands~\cite{wang2024openhands}, and OpenClaw~\cite{openclaw2025} provide extensible frameworks for building custom agent pipelines.

This landscape illustrates a broader shift from conversational competence to operational competence, and it changes where the performance bottleneck lies.
For question answering (QA), incremental improvements in model capability---\ie, larger parameters, more training data, or better alignment---often yield direct and predictable gains.
Yet traditional benchmarks that measure such capability, including
MMLU~\cite{hendrycks2021mmlu}, GPQA~\cite{gpqa}, and
HumanEval~\cite{chen2021humaneval}, have become increasingly saturated at the frontier, with contamination risks further complicating interpretation; harder evaluations such as Humanity's Last Exam~\cite{hle} have been proposed to restore discriminative power.
More critically, when evaluation shifts from closed-form QA to interactive, multi-step task completion, even frontier models reveal substantial reliability gaps---SWE-bench~\cite{jimenez2024swe}, WebArena~\cite{zhou2024webarena}, OSWorld~\cite{xie2025osworld}, TheAgentCompany~\cite{xu2026theagentcompany}, and Terminal-Bench~\cite{merrill2026terminalbench} all demonstrate that agentic tasks retain significant headroom.
This divergence between static benchmarks (nearing saturation) and agentic benchmarks (far from solved) raises a natural question: if model scaling alone does not close the gap on agentic tasks, what does?

\subsection{Harness Design as a Performance Lever}
\label{sec:intro:harness}

A growing body of work suggests that agent performance is increasingly limited not only by the model's raw reasoning power, but also by the design of its \emph{execution harness}: the runtime infrastructure that shapes what the model perceives, how it acts, and whether its errors are detected and recovered.
The idea that interface design matters was first demonstrated empirically by SWE-agent~\cite{yang2024sweagent}, which showed that redesigning the agent--computer interface (ACI) can substantially improve SWE-bench performance under a fixed base model.
The broader concept was subsequently crystallized under the term \emph{harness engineering} by Hashimoto~\cite{hashimoto2026harness} and OpenAI~\cite{openai2026harness}, who framed an agent as \emph{model plus harness} and identified observation shaping, action-space design, execution sandboxing, context management, and verification loops as its core components.
More recently, NLAH~\cite{pan2026nlah} formalizes harness logic as an editable, portable natural-language artifact, and Meta-Harness~\cite{lee2026metaharness} treats harness configuration as an optimizable search space.

Following this line of work, we adopt the harness-centric perspective as a unifying lens: we systematically examine how the design of the runtime infrastructure, rather than model capability alone, determines agent reliability, efficiency, and generalization across diverse tasks.
We further argue that this perspective is now \textbf{extending beyond single-model scaffolds}: recent systems~\cite{opensquilla2026} treat the harness as a compositional runtime over multiple models, and increasingly as a learnable object whose routing, orchestration, and verification policies can themselves be optimized.

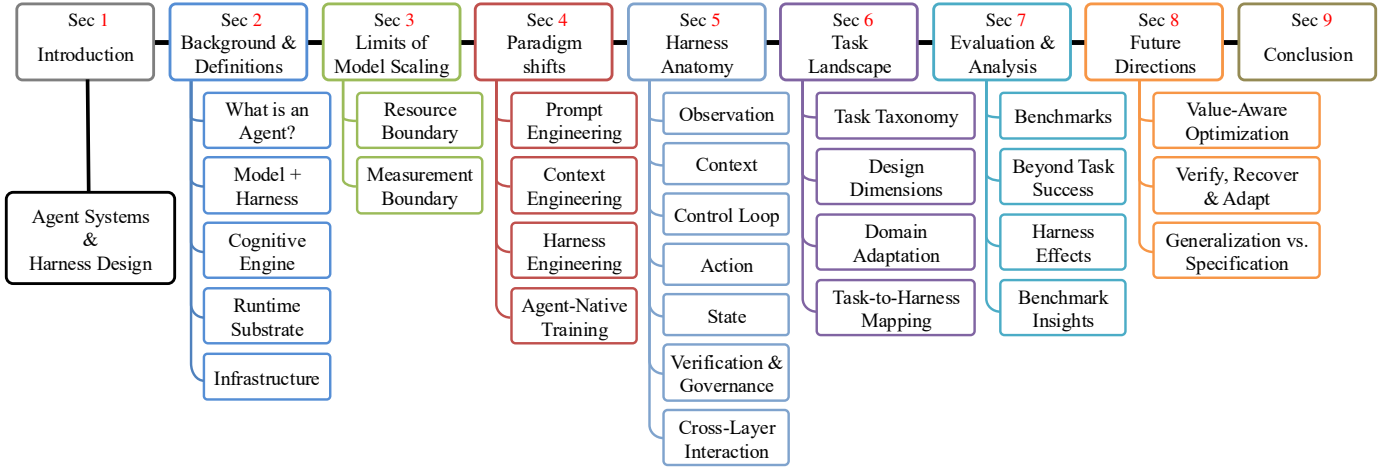
\begin{figure*}[t]
\centering
\begin{overpic}[width=\linewidth]{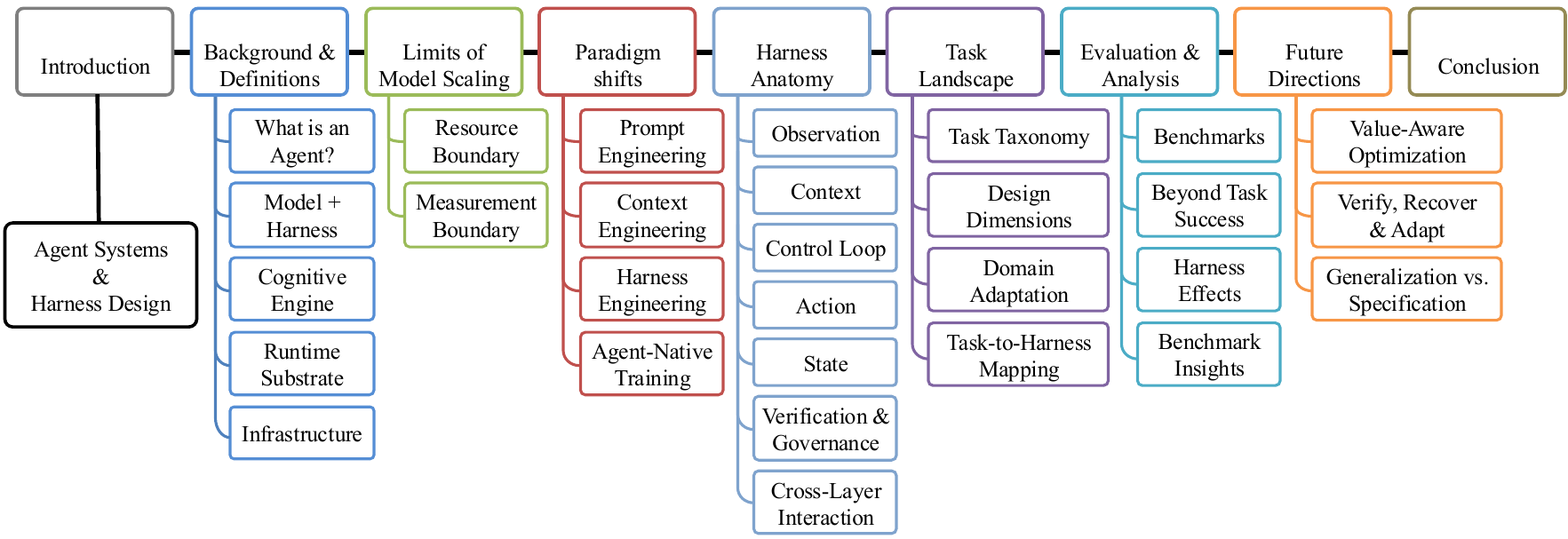}
  \put(4.3,32.2){%
  {\fontfamily{ptm}
   \fontsize{7pt}{11pt}\selectfont
   Sec~\ref{sec:intro}}%
  }
  \put(15.4125,32.2){%
  {\fontfamily{ptm}
   \fontsize{7pt}{11pt}\selectfont
   Sec~\ref{sec:background}}%
  }
  \put(26.525,32.2){%
  {\fontfamily{ptm}
   \fontsize{7pt}{11pt}\selectfont
   Sec~\ref{sec:fm}}%
  }
  \put(37.6375,32.2){%
  {\fontfamily{ptm}
   \fontsize{7pt}{11pt}\selectfont
   Sec~\ref{sec:sec4:path}}%
  }
  \put(48.75,32.2){%
  {\fontfamily{ptm}
   \fontsize{7pt}{11pt}\selectfont
   Sec~\ref{sec:sec5:anatomy}}%
  }
  \put(59.8625,32.2){%
  {\fontfamily{ptm}
   \fontsize{7pt}{11pt}\selectfont
   Sec~\ref{sec:tasks}}%
  }
  \put(70.975,32.2){%
  {\fontfamily{ptm}
   \fontsize{7pt}{11pt}\selectfont
   Sec~\ref{sec:sec7:eval}}%
  }
  \put(82.0875,32.2){%
  {\fontfamily{ptm}
   \fontsize{7pt}{11pt}\selectfont
   Sec~\ref{sec:future}}%
  }
  \put(93.2,32.2){%
  {\fontfamily{ptm}
   \fontsize{7pt}{11pt}\selectfont
   Sec~\ref{sec:conclusion}}%
  }
\end{overpic}
%\vspace{-18pt}
\caption{\small{A diagram that summarizes the structure of this survey.}}
%\vspace{-10pt}
\label{fig:survey-structure}
\end{figure*}

\subsection{Four Paradigms of Agent Engineering}
\label{sec:intro:thesis}

We organize the recent literature through an evolutionary lens of four paradigms.
Each emerged to address limitations exposed by its predecessor; each foregrounds a different performance lever.

\paratitle{\underline{Phase~1: Prompt Engineering}}
optimizes the single-turn instruction sent to the model.
Techniques such as few-shot exemplars~\cite{brown2020gpt3}, chain-of-thought reasoning~\cite{wei2022chain}, self-consistency~\cite{wang2022selfconsistency}, and tree-of-thought search~\cite{yao2023tree} can clarify tasks, constrain output format, and elicit the model's latent capabilities.
Yet prompting fundamentally addresses an \emph{expression} problem: how to ask.
It does not solve the \emph{information} problem: prompting alone cannot \textbf{supply missing knowledge, manage dynamically evolving state, or maintain coherence across long action sequences.}

\paratitle{\underline{Phase~2: Workflows and Context Engineering}}
shifts the unit of optimization from a single prompt to the information lifecycle surrounding multi-step execution.
Its core discipline is curating \emph{what} information enters the model's context window, \emph{when}, and \emph{in what form}~\cite{anthropic2025context}, encompassing retrieval-augmented generation~\cite{lewis2020rag}, long-term memory management~\cite{packer2023memgpt}, tool and API definitions~\cite{schick2023toolformer,patil2023gorilla}, and progressive skill disclosure~\cite{wang2023voyager,xu2026agentskills}.
The evaluation criterion changes accordingly: the question is no longer only whether a single answer is correct, but whether the assembled context enables the model to complete multi-step tasks.
However, context engineering remains fundamentally feedforward: it optimizes the input to each reasoning step but \textbf{provides no structural mechanism to detect drift, verify intermediate outcomes, or recover from errors.}

\paratitle{\underline{Phase~3: Harness Engineering}}
closes the loop.
Beyond assembling the right context, the harness introduces feedback-driven execution: the model acts, observes environment responses, and reasons over observations to decide its next step~\cite{yao2023react}.
More broadly, harness engineering treats the entire runtime infrastructure as the primary design object~\cite{hashimoto2026harness,openai2026harness}, governing execution sandboxing, state checkpointing, verification loops, error recovery, and sub-agent coordination~\cite{pan2026nlah,lee2026metaharness}.
The governing question shifts from \emph{what to show the model} to \emph{how to keep the whole system on track}: \textbf{prevent drift, maintain stable execution, and recover from errors.}

Within this phase, a further shift is already visible.
Early harness design typically wraps a \emph{fixed} foundation model with hand-crafted or searched runtime policies.
More recent systems move toward a \emph{multi-model harness}: the runtime routes, delegates, and composes heterogeneous models for planning, tool use, verification, coding, and domain-specific subtasks~\cite{fourney2024magenticone,zhu2026symphony,openai2025agentssdk}.
At the same time, the harness itself is becoming \emph{learnable}: harness modules, orchestration logic, and runtime policies can be edited, searched, or optimized as first-class artifacts~\cite{pan2026nlah,lee2026metaharness,lin2026agenticHarnessEngineering}.
This changes what counts as an agent system.
A single prompt wrapped around one model can still function as a lightweight agent, but reliable long-horizon task completion increasingly depends on \textbf{a compositional, optimizable runtime over multiple models}, not on prompt craft alone.

\paratitle{\underline{Phase~4: Agent-Native Training and Co-Evolution}}
builds on the learnable multi-model harness view above.
Its first direction is \emph{internalization}: agentic behaviors such as planning, tool use, verification, and recovery are increasingly trained into model parameters through interactive environments~\cite{qi2024webrl,lai2025computerrl,deepseek2025r1,wu2025evolver}.
Its second direction is \emph{co-evolution}: over deployment, the model, harness, and improvement loop may all be updated from execution traces that indicate what to keep, change, or undo~\cite{wu2025evolver,zhai2025agentevolver,zhang2025darwin,lin2026agenticHarnessEngineering}.
This does not eliminate the harness; it shifts the design question toward how much of agent behavior is learned in models, how much stays in the runtime, and how the full stack improves safely over time, opening a path toward \textbf{self-evolving agent systems}.

These four phases form a conceptual evolutionary lens rather than a strict temporal partition; all four coexist in practice today.
Our goal is not to introduce another component taxonomy, but to use this progression and benchmark evidence (Sec.~\ref{sec:sec7:eval}) to analyze how the dominant performance bottleneck moves across stages, and why harness design has become a central object of agent engineering.

\begin{table*}[t]
\centering
\caption{\small{Comparison between our work and representative prior surveys. 
``Broad'' denotes coverage of the general LLM-based agent landscape rather than a specific subfield; 
``Eval.'' denotes explicit coverage of benchmarks and the evaluation of methods;
``App.'' denotes substantial discussion of application domains and use cases; 
and ``Industry'' denotes the extent to which a survey incorporates practitioner reports, production systems, or industrial engineering evidence as part of its main analysis.
% Time indicates the first public release of each survey.
}}
\label{tab:survey-comparison}
\footnotesize
\setlength{\aboverulesep}{0pt}
\setlength{\belowrulesep}{0pt}
\renewcommand{\arraystretch}{1.0}
\begin{tabular}
{m{1.96cm}>{\centering\arraybackslash}m{0.8cm}>{\centering\arraybackslash}m{2.34cm}m{7.1cm}>{\centering\arraybackslash}m{0.56cm}>{\centering\arraybackslash}m{0.56cm}>{\centering\arraybackslash}m{0.56cm}>{\centering\arraybackslash}m{1.0cm}}
\toprule
\textbf{Survey} & \textbf{Time} & \textbf{Organizing lens} & \textbf{Primary focus} & \textbf{Broad} & \textbf{Eval.} & \textbf{App.} & \textbf{Industry} \\
\midrule
Wang \etal~\cite{wang202wang2024autonomoussurvey4survey} & 2023.08 & Module-based agent construction & How to construct an autonomous LLM agent through core modules, \eg, profile, memory, planning, and action. & \cmark & \xmark & \cmark & No \\
\rowcolor{gray!15}
Xi \etal~\cite{xi2025rise} & 2023.09 & Brain--perception--action framework & Agents as intelligent systems, from single-agent arch. to multi-agent society and human--agent interaction. & \cmark & \xmark & \cmark & No \\
Luo \etal~\cite{luo2025llmagent} & 2025.03 & Build--collaborate--evolve taxonomy & Taxonomy of agents spanning methodological foundations, collaboration, applications, and evaluation. & \cmark & \cmark & \cmark & Limited \\
\midrule
\rowcolor{gray!15}
Guo \etal~\cite{guo2024llmbased} & 2024.02 & Communication and collaboration & Overall progress, communication patterns, and open challenges in LLM-based multi-agent systems. & \xmark & \xmark & \cmark & No \\
Li \etal~\cite{li2024survey} & 2024.10 & Workflow-based taxonomy & How multi-agent systems are structured through workflow, infrastructure, core functional modules. & \xmark & \xmark & \xmark & Limited \\
\rowcolor{gray!15}
Li \etal~\cite{li2025collaboration} & 2025.01 & Collaboration mechanism & Collaboration in multi-agent systems, categorized by actors, structures, strategies and coordination protocols. & \xmark & \xmark & \cmark & No \\
\midrule
Shen \etal~\cite{yehudai2025survey} & 2025.03 & Evaluation-based taxonomy & Benchmarks, metrics, and methodological issues in evaluating LLM-agents. & \xmark & \cmark & \xmark & Limited \\
\midrule
\rowcolor{gray!15}
Gu \etal~\cite{gu2025guiagents} & 2025.07 & Domain-focused taxonomy & GUI/computer-use agents: benchmarks, architectures, and training methods. & \xmark & \cmark & \cmark & Limited \\
Ma \etal~\cite{ma2024vla} & 2024.05 & Embodied-agent taxonomy & Vision-language-action models for embodied AI. & \xmark & \cmark & \cmark & No \\
\rowcolor{gray!15}
Zhang \etal~\cite{zhang2025trustworthy} & 2025.03 & Safety-oriented taxonomy & Threats, safety risks, evaluation, and countermeasures for trustworthy LLM-based agents. & \xmark & \cmark & \cmark & Limited \\
\midrule
Meng \etal~\cite{meng2026agent} & 2026.04 & Execution harness taxonomy & Six-component tuple for harness definition, historical tracing, and cross-cutting harness challenges. & \cmark & \xmark & \xmark & Strong \\
\rowcolor{gray!15}
Li \etal~\cite{li2026agent} & 2026.04 & Seven-layer harness taxonomy & ETCLOVG, a seven-layer taxonomy and practitioner principles from deployed agent stacks. & \cmark & \xmark & \cmark & Strong \\
Ning \etal~\cite{ning2026code} & 2026.05 & Code-as-harness layers & Code as executable harness substrate: interface and multi-agent scaling across application domains. & \xmark & \cmark & \cmark & Limited \\
%\midrule
\rowcolor{gray!12}
\textbf{Ours} & \textbf{2026.06} & \textbf{Engineering paradigm shifts} & \textbf{How agent engineering evolved from prompt optimization to runtime system design and future directions.} & \textbf{\cmark} & \textbf{\cmark} & \textbf{\cmark} & \textbf{Strong} \\
\bottomrule
\end{tabular}
\end{table*}

\subsection{Relation to Prior Surveys}
\label{sec:intro:related}

Recent surveys have documented the rapid rise of LLM-based agents, but most organize the field through taxonomy-oriented lenses.
Tab.~\ref{tab:survey-comparison} compares our survey with representative prior work.
General-purpose surveys summarize agent architectures and components such as memory, planning, action, perception, applications, safety, and evaluation~\cite{wang202wang2024autonomoussurvey4survey,xi2025rise,luo2025llmagent}.
Multi-agent surveys focus on communication, coordination, collaboration structures, and workflow organization~\cite{guo2024llmbased,li2024survey,li2025collaboration}.
Other surveys examine narrower but important slices, including evaluation methodology~\cite{yehudai2025survey}, GUI/computer-use agents~\cite{gu2025guiagents}, embodied systems~\cite{ma2024vla}, and trustworthy agents~\cite{zhang2025trustworthy}.
Since early 2026, several works have narrowed the lens specifically to agent harnesses.
Meng \etal~\cite{meng2026agent} formalize the harness as a six-component tuple.
Li \etal~\cite{li2026agent} further propose the seven-layer ETCLOVG taxonomy and map a large open-source corpus onto it to expose ecosystem coverage and production design principles.
Ning \etal~\cite{ning2026codeasharness} organize the field from a code-centric perspective, treating executable programs as the substrate for reasoning, action, state, and verification.
These surveys provide valuable harness taxonomies, catalogs, or substrate-specific roadmaps.

Relative to recent harness-focused surveys, our contribution is not primarily another layer taxonomy or project catalog.
We instead ask how the dominant engineering bottleneck migrates across prompt optimization, context/workflow organization, compositional and learnable runtimes, and agent-native co-evolution, and how that migration should be evaluated empirically.
Accordingly, we connect harness anatomy to task pressure profiles (Sec.~\ref{sec:tasks}), benchmark evidence (Sec.~\ref{sec:sec7:eval}), and value-aware deployment objectives (Sec.~\ref{sec:future}), rather than centering the analysis on taxonomy completeness or repository coding alone.

Our survey makes three distinctions explicit.
First, it is \emph{evolution-first}: we organize the literature around engineering paradigm shifts rather than a static component taxonomy.
Second, it is \emph{harness-centric}: we treat the execution harness as a first-class technical object that governs observation, context, control, action, state, verification, recovery, and efficiency.
Third, it connects \emph{academic evidence with industrial practice}, using benchmark results, open-source systems, engineering reports, and controlled model--harness analyses to examine how runtime design choices affect agent reliability, cost, and latency.

In short, our goal is not only to catalog LLM-based agents, but to explain why harness engineering emerged as a central systems concern and how it may extend toward future agent-native training and co-evolution.

\subsection{Scope Boundaries}
\label{sec:intro:scope}

This survey covers LLM-based agent systems from 2020 to 2026, including prompting methods, workflow frameworks, harness and runtime design, multi-model orchestration, agent-native training, model--harness co-evolution, domain deployments, and evaluation methodology.
We focus on systems in which one or more LLMs serve as cognitive engines within an execution harness, and synthesize published papers, public engineering reports, benchmarks, and controlled model--harness comparisons.
We prioritize high-impact and verifiable sources that directly inform agent system design, efficiency, or evaluation.
Adjacent traditions such as neuro-symbolic planning, classical embodied control, and non-LLM systems are treated as complementary work rather than surveyed in depth, because they rely on different assumptions, architectures, and evaluation criteria.

\subsection{Contributions and Survey Structure}
\label{sec:intro:contributions}

The main contributions of this survey are:

\begin{itemize}[leftmargin=*]
  \item We provide an evolution-first synthesis of agent engineering, tracing shifts from prompt engineering to context engineering, harness engineering, and agent-native training.
  \item We analyze the limits of model-centric scaling for long-horizon task completion and argue that agent performance is a property of the model--harness pairing.
  \item We formalize the execution harness as a runtime design object and decompose it into six coupled responsibilities.
  \item We map task properties, domain adaptations, and evaluation practices to harness pressure profiles rather than treating agent components as an independent checklist.
  \item We synthesize benchmark and empirical evidence to motivate value-aware evaluation beyond task success.
\end{itemize}

The remainder of this survey is organized as follows.
Sec.~\ref{sec:background} defines agents and harnesses and reviews core infrastructure.
Sec.~\ref{sec:fm} analyzes the limits of model-centric scaling.
Sec.~\ref{sec:sec4:path} presents the four-paradigm evolution of agent engineering.
Sec.~\ref{sec:sec5:anatomy} decomposes the execution harness into six runtime components.
Sec.~\ref{sec:tasks} maps task pressures and domain adaptations to harness configurations.
Sec.~\ref{sec:sec7:eval} reviews benchmarks, evaluation methodology, and model--harness evidence.
Sec.~\ref{sec:future} discusses open challenges and future directions, and Sec.~\ref{sec:conclusion} concludes.
Fig.~\ref{fig:survey-structure} summarizes the overall structure of this survey.

\section{Background and Definitions}

\label{sec:background}
Two abstraction levels are often conflated in the agent literature.
At the functional level, an agent is a goal-directed closed-loop system: it perceives an external environment, maintains task state, reasons and decides, executes actions, and adapts from feedback.
At the implementation level, an LLM-based agent is not the foundation model alone, but a coupled system consisting of a foundation model and an execution harness.
The model supplies flexible language understanding, reasoning, planning, and action proposal; the harness supplies the runtime machinery that exposes observations, constructs context, executes actions, persists state, and verifies or recovers from failures.
This distinction reconciles classical agent definitions with recent harness-centered accounts of LLM agents: the former define \emph{what} an agent must do, whereas the latter specify \emph{how} those functions are realized in deployed systems.

\begin{figure}[t]
\centering
\includegraphics[width=\linewidth]{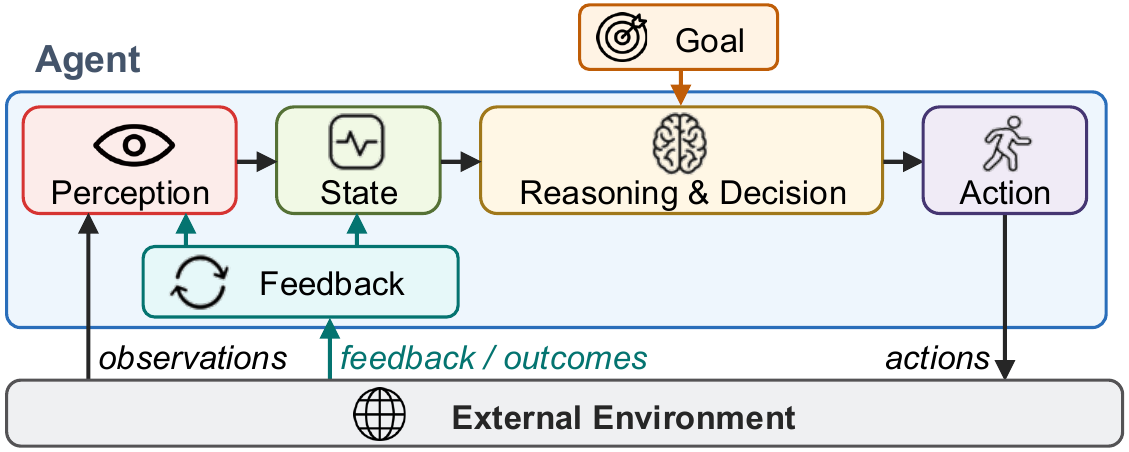}
\caption{\small{Functional view of an agent: a goal-directed closed-loop system that receives observations from external environments, maintains state, reasons and acts on the environment, and adapts from feedback or outcomes. This view defines \emph{what} an agent must do, independent of any particular implementation.}}
\label{fig:sec2:agent-concept}
\end{figure}

\subsection{Functional View: What Is an Agent?}
The notion of an agent predates LLMs. Wooldridge and Jennings~\cite{wooldridge1995intelligent} characterize an intelligent agent as a system situated in an environment, able to perceive that environment and act upon it in pursuit of goals~\cite{park2023generative,wu2023autogen,hong2024metagpt}.
For this survey, the defining property is not whether the system is implemented by symbolic rules, reinforcement learning, or a language model, but whether it sustains a goal-conditioned loop with its environment.
We therefore use a functional definition: an agent is a system that organizes five operations around a task objective:
\textbf{perception}, \textbf{state maintenance}, \textbf{reasoning and decision-making}, \textbf{action}, and \textbf{feedback adaptation}.
The goal and environment condition a particular run, but they are not themselves internal components of the agent.
As illustrated in Fig.~\ref{fig:sec2:agent-concept}, the agent receives observations, updates internal or external state, chooses the next action through reasoning and decision-making, acts on the environment, and incorporates feedback or outcomes into subsequent behavior.
This closed-loop property separates agents from single-shot model calls, static retrieval systems, and fixed automation scripts that do not revise behavior as observations change.

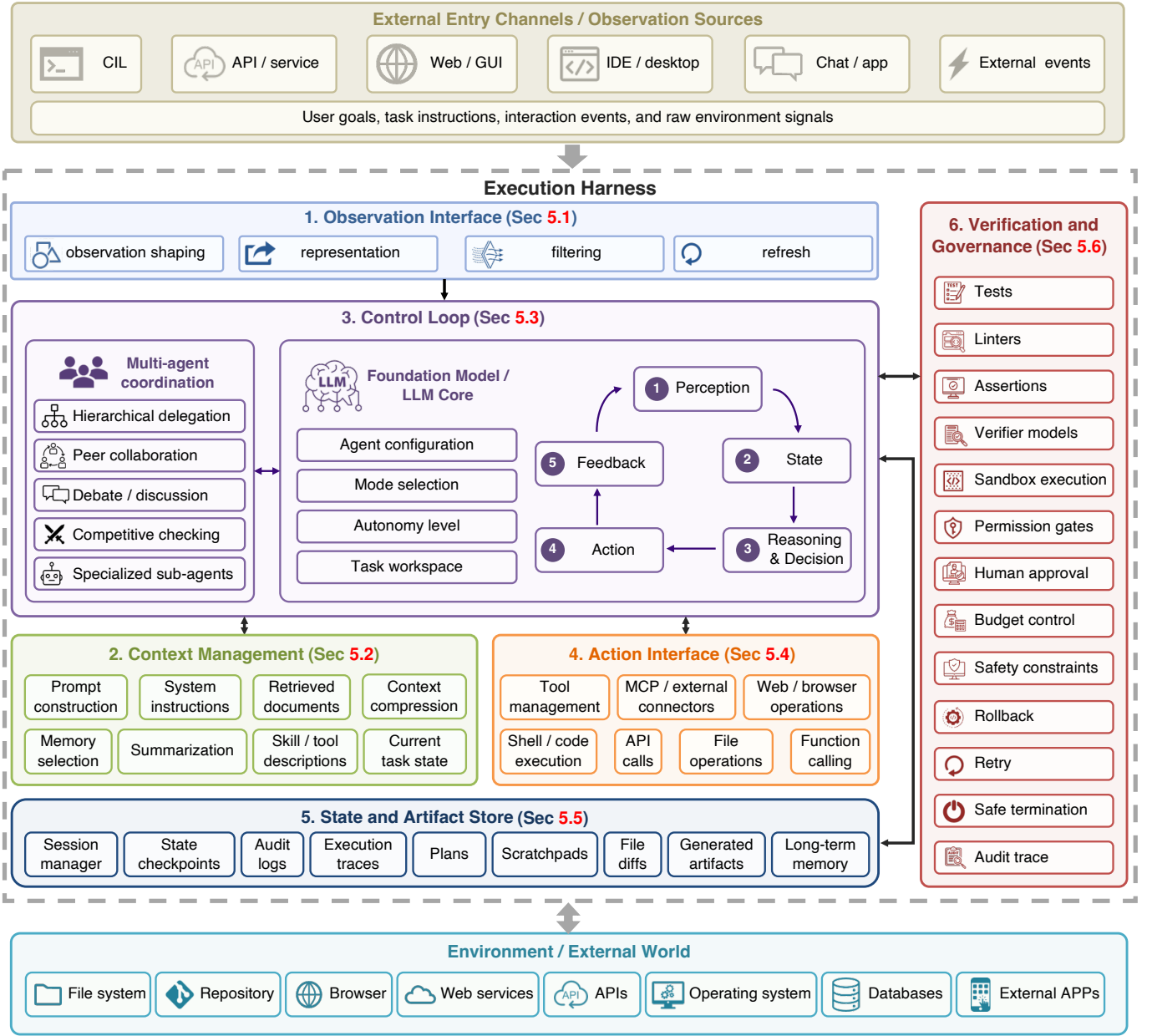
\begin{figure*}[!ht]
\centering
\begin{overpic}[width=\linewidth]{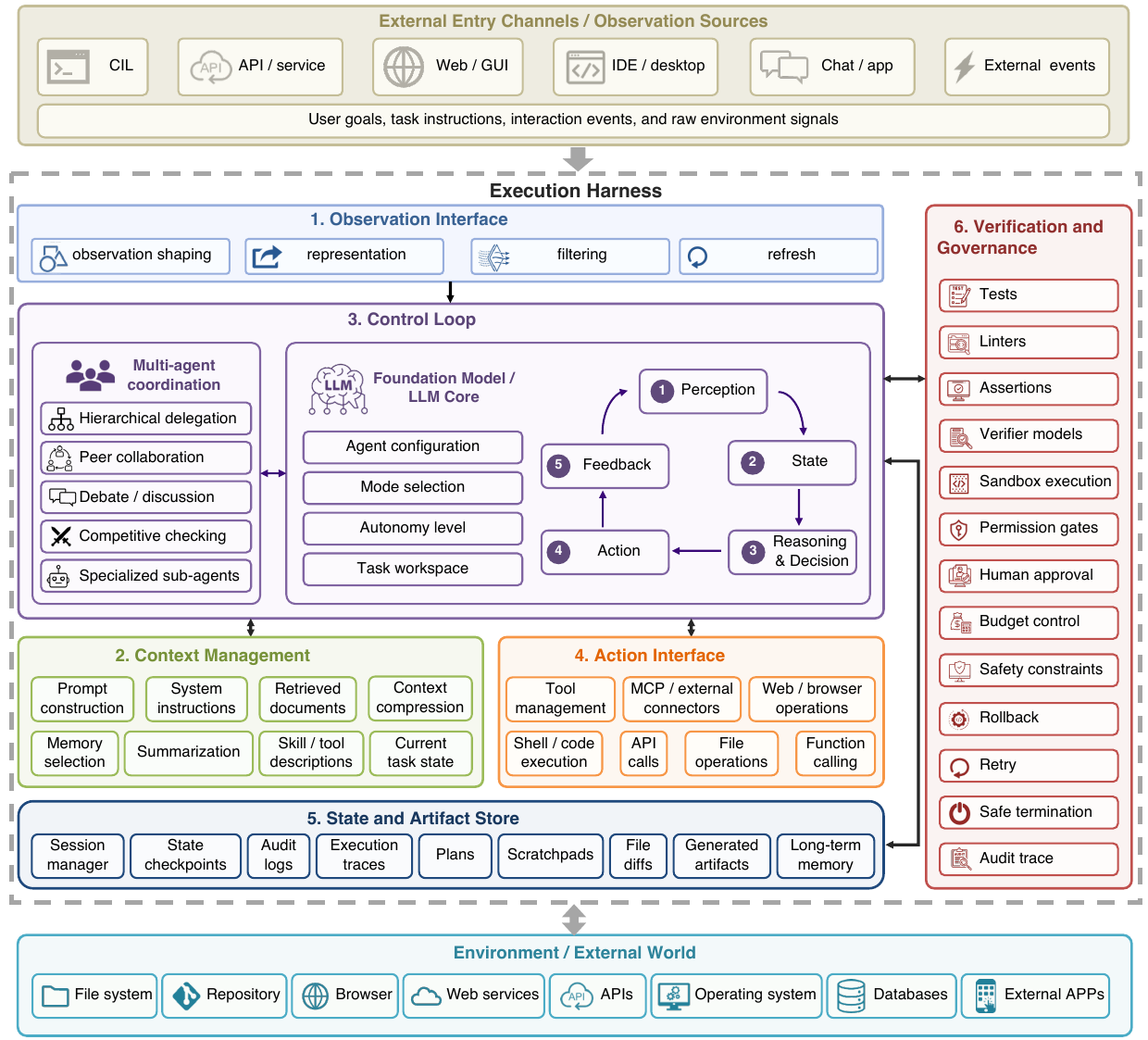}
  \put(44.6,70.9){%
  {\fontfamily{phv}\bfseries
   \fontsize{7.8pt}{11pt}\selectfont
   \textcolor[HTML]{386193}{(Sec~\ref{sec:anatomy-observation-interface})}}%
  }
  \put(27.4,32.95){%
  {\fontfamily{phv}\bfseries
   \fontsize{7.8pt}{11pt}\selectfont
   \textcolor[HTML]{769138}{(Sec~\ref{sec:anatomy-context-manager})}}%
  }
  \put(41.8,62.2){%
  {\fontfamily{phv}\bfseries
   \fontsize{7.8pt}{11pt}\selectfont
   \textcolor[HTML]{5B4477}{(Sec~\ref{sec:anatomy-control-loop})}}%
  }
  \put(63.6,32.95){%
  {\fontfamily{phv}\bfseries
   \fontsize{7.8pt}{11pt}\selectfont
   \textcolor[HTML]{E36400}{(Sec~\ref{sec:anatomy-action-interface})}}%
  }
  \put(45.6,18.7){%
  {\fontfamily{phv}\bfseries
   \fontsize{7.8pt}{11pt}\selectfont
   \textcolor[HTML]{17375E}{(Sec~\ref{sec:anatomy-state-artifact-store})}}%
  }
  \put(90.7,68.4){%
  {\fontfamily{phv}\bfseries
   \fontsize{7.8pt}{11pt}\selectfont
   \textcolor[HTML]{902E2D}{(Sec~\ref{sec:anatomy-verification-governance})}}%
  }
\end{overpic}
\caption{\small{Implementation view of an LLM-based agent as a foundation model coupled with an execution harness. The harness mediates closed-loop interaction between the model and the external world through six runtime components: observation interface, context manager, control loop, action interface, state and artifact store, and verification and governance layer. The section labels inside the figure indicate where each component is analyzed in detail.}}
\label{fig:sec2:agent-implementation}
\end{figure*}

\subsection{Implementation View: Model Plus Harness}
In LLM-based agents, the functional loop is implemented by more than an inference call to a model.
The foundation model is necessary because it provides the general-purpose cognitive capabilities that make open-ended task completion possible.
It is not sufficient, however, because the model does not by itself define what observations are available, which actions are permitted, where long-term state is stored, how execution is validated, or how failures are recovered.
Following recent industrial and academic discussions of harness engineering~\cite{hashimoto2026harness,openai2026harness}, we write LLM-based agent as:
\begin{equation}
\mathcal{A}_{\mathrm{LLM}}
=
\langle \mathcal{M}, \mathcal{H} \rangle
=
\langle
\mathcal{M},
\mathcal{I}_{\mathrm{obs}},
\mathcal{C},
\mathcal{L},
\mathcal{I}_{\mathrm{act}},
\mathcal{S},
\mathcal{V}
\rangle ,
\end{equation}

\noindent where $\mathcal{M}$ denotes the model layer of the agent.
In the simplest case, $\mathcal{M}$ is a single foundation model.
In deployed multi-model systems, $\mathcal{M}=\{\mathcal{M}_1,\ldots,\mathcal{M}_k\}$ denotes a set of backbone models with heterogeneous capabilities, costs, and context limits~\cite{fourney2024magenticone,zhu2026symphony,openai2025agentssdk}.
$\mathcal{H}$ denotes the execution harness surrounding $\mathcal{M}$.
The second equality expands the harness into six runtime components used throughout this survey.
This expression is an implementation-oriented decomposition, not a replacement for the functional definition above.
When $|\mathcal{M}|=1$, the agent reduces to the familiar single-backbone setting; when $|\mathcal{M}|>1$, the harness must additionally decide \emph{which} model acts at each step~\cite{opensquilla2026}.
The model layer and harness jointly instantiate the functional loop: the active model reasons over a supplied context and proposes next steps, while the harness determines what it sees, what it can do, how execution state persists, and how errors are detected, constrained, or repaired.

\subsection{LLM as the Cognitive Engine}
LLMs became viable cognitive engines for agents because they combine capabilities that previously required separate modules or task-specific policies.

\paratitle{Reasoning and planning.}
Prompting methods such as chain-of-thought~\cite{wei2022chain}, Tree of Thoughts~\cite{yao2023tree,long2023tot}, self-consistency~\cite{wang2022selfconsistency}, and Reflexion~\cite{shinn2023reflexion} show that sufficiently capable models can support task decomposition, branching search, self-critique, and multi-step inference.
These abilities make the model a plausible decision engine for tasks whose solution cannot be enumerated in advance.

\paratitle{In-context adaptation.}
The same frozen model can adapt its behavior through instructions, examples, retrieved documents, tool descriptions, and intermediate artifacts.
This reduces the need to train a separate policy for each environment, while making the quality, ordering, and compression of the supplied context a primary determinant of behavior.

\paratitle{Action proposal and tool use.}
When models can emit structured tool calls~\cite{schick2023toolformer,patil2023gorilla}, they are no longer limited to internal text generation.
They can propose calls to code execution, retrieval systems, browsers, APIs, and external software.
Yet a proposal is not an executed action: reliability depends on the harness to validate, dispatch, observe, and, when necessary, reject or repair the proposed action.

These strengths also expose the model's limitations.
LLMs remain vulnerable to hallucination, finite context windows, weak persistent memory, prompt sensitivity, and limited intrinsic ability to verify long-horizon outcomes.
The harness is therefore not an optional engineering wrapper; it is the runtime layer that turns model capability into sustained, inspectable interaction with an environment.

\subsection{Harness as the Runtime Substrate}
\label{sec:sec2.4-harness}
Following~\cite{hashimoto2026harness,openai2026harness,anthropic2025harness}, we use \emph{harness} to denote the runtime infrastructure that surrounds the model and realizes closed-loop agent execution.
The harness is broader than an individual tool, memory module, prompt template, or workflow script.
It is the coordinating layer that decides which observations reach the model, how context is assembled, how the agent loop advances, how actions are executed, how state and artifacts persist, and how failures are detected, governed, and recovered.
This runtime substrate can be formalized as:
\begin{equation}
\mathcal{H}
=
\langle
\mathcal{I}_{\mathrm{obs}},
\mathcal{C},
\mathcal{L},
\mathcal{I}_{\mathrm{act}},
\mathcal{S},
\mathcal{V}
\rangle .
\end{equation}

\noindent The six components are:

\begin{itemize}[leftmargin=*]
  \item \textbf{Observation interface}~$\mathcal{I}_{\mathrm{obs}}$: transforms raw environment signals into model-usable observations, including terminal output, file diffs, screenshots, DOM states, API responses, logs, retrieved passages, and event streams.
  \item \textbf{Context manager} $\mathcal{C}$: determines what information enters the model context, when it enters, and in what form, covering prompt construction, system instructions, retrieval, memory selection, compression, summarization, tool descriptions, and current task state.
  \item \textbf{Control loop} $\mathcal{L}$: orchestrates the observe-reason-act-feedback cycle, including step scheduling, stopping criteria, retries, reflection, delegation, handoffs, and multi-agent coordination. In multi-model settings, $\mathcal{L}$ additionally implements model routing and role assignment.
  \item \textbf{Action interface} $\mathcal{I}_{\mathrm{act}}$: maps model outputs to executable operations, such as function calls, MCP tools, shell or code execution, browser actions, file operations, API calls, and sub-agent invocations.
  \item \textbf{State and artifact store} $\mathcal{S}$: persists execution state and products, including conversation history, plans, scratchpads, checkpoints, logs, traces, diffs, memory records, generated files, and task artifacts.
  \item \textbf{Verification and governance layer} $\mathcal{V}$: checks, constrains, and repairs execution through tests, assertions, verifier models, sandbox policies, permission gates, rollback, retry, budget control, safety constraints, and audit traces.
\end{itemize}

Fig.~\ref{fig:sec2:agent-implementation} visualizes this implementation view.
The figure should be read as an execution architecture rather than a static checklist: observations, context, control, actions, state, and verification form a coupled runtime around the model, and their interaction determines whether model capability becomes reliable task completion.

This decomposition differs from earlier component taxonomies because it is operational rather than purely functional.
For example, memory appears in the functional loop as state, but in a deployed system it may be realized through context selection, artifact storage, retrieval indices, session managers, or checkpointing policies.
Similarly, action is not merely an abstract action space; it is mediated by schemas, permissions, sandboxes, execution APIs, and side-effect controls.
By separating these runtime responsibilities from the model itself, the decomposition explains why harness changes can improve agent performance even when the underlying model is unchanged~\cite{yang2024sweagent,pan2026nlah,lee2026metaharness}.

\subsection{Key Infrastructure Primitives}
Several infrastructure primitives recur across modern LLM-based agents.
They should not be treated as concepts parallel to the harness.
Rather, they instantiate specific harness responsibilities in deployed systems and make perception, action, communication, and governance concrete.

\paratitle{Tool and function calling.}
Structured tool invocation converts model outputs from free-form suggestions into machine-executable calls~\cite{schick2023toolformer,patil2023gorilla}.
Tool schemas are primarily part of the action interface $\mathcal{I}_{\mathrm{act}}$, while tool descriptions, arguments, and returned results also shape the context manager $\mathcal{C}$ and observation interface $\mathcal{I}_{\mathrm{obs}}$.

\paratitle{Model Context Protocol (MCP).}
MCP~\cite{mcp2025} standardizes how LLM applications expose tools, data sources, and contextual resources to agents.
In our notation, MCP primarily strengthens the boundary between the context manager and action interface by reducing connector fragmentation and making tool and data access more modular.

\paratitle{Agent-to-Agent communication.}
The Agent2Agent (A2A) protocol~\cite{google2025a2a} targets interoperability among agents built by different vendors or frameworks.
It is most relevant to the control loop $\mathcal{L}$ and action interface $\mathcal{I}_{\mathrm{act}}$, especially when delegation, negotiation, debate, or multi-agent collaboration becomes part of the execution process.

\paratitle{Sandboxed execution and approval.}
When agents can write files, execute code, browse the web, or call APIs, isolation becomes both a safety mechanism and a reproducibility primitive.
Sandboxes constrain filesystem access, network egress, process execution, and resource usage, while approval policies determine when human authorization is required before an action is dispatched.
These mechanisms belong primarily to the verification and governance layer $\mathcal{V}$.

\paratitle{Agent SDKs and tracing.}
Frameworks such as the OpenAI Agents SDK~\cite{openai2025agentssdk} expose reusable abstractions for tools, handoffs, tracing, and loops.
They package common harness patterns into developer-facing interfaces, making runtime behavior more reusable, inspectable, and debuggable.

Tab.~\ref{tab:function-harness-map} summarizes how the conceptual operations in the functional agent loop are realized by the implementation components defined above.
The mapping is many-to-many rather than one-to-one: perception depends not only on the observation interface, but also on the context manager that selects and formats observations for the model; feedback involves verification, control-loop decisions, and state updates.
This many-to-many mapping bridges the conceptual view in Fig.~\ref{fig:sec2:agent-concept} and the implementation view in Fig.~\ref{fig:sec2:agent-implementation}.

\begin{table}[!t]
\centering
\caption{\small{Mapping from conceptual agent operations to harness realizations.}}
\label{tab:function-harness-map}
\small
\setlength{\aboverulesep}{0pt}
\setlength{\belowrulesep}{0pt}
\setlength{\tabcolsep}{4pt}
\begin{tabular}{>{\centering\arraybackslash}m{1.7cm}
>{\centering\arraybackslash}m{1.65cm}
>{\centering\arraybackslash}m{4.5cm}}
\toprule
\addlinespace[2pt]
\textbf{\shortstack[c]{Functional\\operation}} 
& \textbf{\shortstack[c]{Harness\\components}} 
& \textbf{Typical mechanisms} \\
\midrule
\rowcolor{gray!15}
Perception & $\mathcal{I}_{\mathrm{obs}}, \mathcal{C}$ & Logs, DOMs, screenshots, retrieval, summaries \\
State Maintenance & $\mathcal{S}, \mathcal{C}$ & Memory, checkpoints, artifacts, conversation history \\
\rowcolor{gray!15}
Reasoning, Decision & $\mathcal{M}, \mathcal{C}, \mathcal{L}$ & Prompted reasoning, plans, tool-choice context \\
Action & $\mathcal{I}_{\mathrm{act}}, \mathcal{V}$ & Function calls, shell commands, APIs, approval gates \\
\rowcolor{gray!15}
Feedback Adaptation & $\mathcal{V}, \mathcal{L}, \mathcal{S}$ & Tests, reflection, retries, rollback, trace updates \\
\bottomrule
\end{tabular}
\end{table}

With these definitions in place, common application labels can be read as specializations of the same model-harness architecture.
Coding, web/GUI, research, embodied, and domain-specific agents all rely on the same six runtime components, but they stress different parts of the harness because their observation channels, action spaces, feedback signals, and safety constraints differ.
Representative examples are summarized in Tab.~\ref{tab:agent-taxonomy}.

\begin{table}[t]
\centering
\caption{\small{Representative types of LLM-based agents.}}
\label{tab:agent-taxonomy}
\small
\setlength{\aboverulesep}{0pt}
\setlength{\belowrulesep}{0pt}
\setlength{\tabcolsep}{2.5pt}
\begin{tabular}{>{\centering\arraybackslash}m{1.5cm}
>{\centering\arraybackslash}m{2.2cm}
>{\centering\arraybackslash}m{1.9cm}
>{\centering\arraybackslash}m{2.5cm}}
\toprule
\addlinespace[2pt]
\textbf{Type} 
& \textbf{Examples} 
& \textbf{Environment} 
& \textbf{Challenge} \\
\midrule
\rowcolor{gray!15}
Coding & Claude Code, Codex & Repository, terminal & Long-horizon reliability \\
Web / GUI & Operator, VisualWebArena & Browser, desktop & Grounding, safe interaction \\
\rowcolor{gray!15}
Research & Deep Research, Elicit & Web, literature & Synthesis, citation fidelity \\
Embodied & Voyager & Real world & Sim-to-real transfer, safety \\
\rowcolor{gray!15}
Domain-specific & ChemCrow, Agent Hospital & Specialized tools & Compliance, domain expertise \\
\bottomrule
\end{tabular}
\end{table}

\section{The Limits of Model-Centric Scaling}
\label{sec:fm}

Once an LLM-based agent is viewed as a model coupled with an execution harness, the role of foundation-model scaling can be stated more precisely.
Scaling remains one of the main reasons why LLMs can serve as the cognitive engine of modern agents.
Scaling laws first showed that language-modeling loss improves predictably with model size, data, and compute~\cite{kaplan2020scaling}, while Chinchilla-style results refined this picture by emphasizing compute-optimal allocation between model parameters and training tokens~\cite{hoffmann2022training}.
The empirical impact is broad: larger and better-trained models have improved reasoning and problem solving~\cite{chowdhery2023palm}, code generation~\cite{nijkamp2022codegen,li2022competition}, mathematical reasoning~\cite{lewkowycz2022solving,shao2024deepseekmath}, and multimodal understanding~\cite{alayrac2022flamingo,chen2024scaling}.
These gains make stronger foundation models indispensable to agent systems, but they don't make model size a complete account of agent performance.
Long-horizon task completion is a trajectory-level property: an agent must repeatedly observe, construct context, choose actions, preserve state, interpret feedback, and recover from errors.
The relevant question is therefore where model-centric explanation stops and runtime design begins.
Two boundaries are especially important: a \emph{resource-performance boundary} and a \emph{measurement boundary}.

\subsection{Resource-Performance Boundary}

The first limit concerns how much additional capability is obtained for additional resources.
Increasing model capacity continues to improve frontier performance, but the gains are increasingly costly, uneven across capabilities, and constrained by inference latency and deployment complexity.
LLaMA3.1~\cite{llama3.1} provides a representative example: moving from the 70B model to the 405B model increased training compute from 7.0M to 30.84M H100 GPU hours, but yielded modest gains on several representative benchmarks, including 2.6 points on MMLU~\cite{hendrycks2021mmlu}, 2.6 points on MBPP EvalPlus~\cite{EvalPlus}, and 1.7 points on GSM8K~\cite{cobbe2021training}.
The larger model also remained far from near-perfect performance on harder reasoning benchmarks such as MATH~\cite{MATH} and MMLU-Pro~\cite{mmlu-pro}.
Similar uneven returns are also visible in Qwen3~\cite{qwen3}, where gains from a much larger base model vary substantially across benchmarks.

\begin{figure}[t]
\centering
\includegraphics[width=\linewidth]{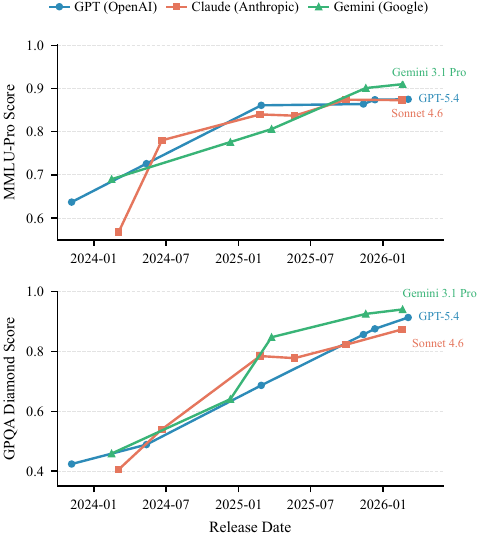}
\caption{\small{
Evolution of frontier-model performance on MMLU-Pro and GPQA Diamond.
Recent GPT, Claude, and Gemini releases increasingly occupy a narrow high-score range on both benchmarks, making later improvements less discriminative than earlier model-generation jumps.
}}
\label{fig:frontier-benchmark-evolution}
\end{figure}

Closed-source frontier models show the same pattern at the high-performance end.
MMLU-Pro~\cite{mmlu-pro,mmlu-pro-leaderboard} and GPQA Diamond~\cite{gpqa,epoch-gpqa-diamond} remain challenging, but recent GPT, Claude, and Gemini releases increasingly cluster within a narrow score range, as shown in Fig.~\ref{fig:frontier-benchmark-evolution}.
This does not mean that scaling has stopped working.
Rather, once models enter a high-accuracy regime on common evaluations, additional scale often yields smaller and more capability-specific improvements while imposing higher cost, latency, and operational burden.
For agents, this trade-off is amplified: a deployed agent invokes the model repeatedly across an execution trajectory, so per-call cost and latency accumulate, and small errors can compound over many steps.

\subsection{Measurement Boundary}

The second limit concerns how scaling-driven progress is measured.
Model-centric progress has often been validated through aggregate gains on static benchmarks.
This was informative when benchmarks clearly separated model generations, but it becomes less discriminative when frontier systems cluster near the upper range of same metrics.
The compression in Fig.~\ref{fig:frontier-benchmark-evolution} illustrates the problem: small score differences on saturated benchmarks are difficult to interpret as meaningful differences in real-world agent capability.

The deeper issue is structural.
Many traditional benchmarks are static, short-horizon, and self-contained: the input is fixed, the output is evaluated once, and the environment does not change in response to the model's actions.
Agent tasks have a different form,  they require long-context understanding, multi-step reasoning, environment interaction, tool use, adaptation to underspecified goals, and robustness to intermediate errors~\cite{swe_agi_benchmarking_specification_driven_2602_09447,agent_diff_benchmarking_llm_agents_2602_11224,featurebench_benchmarking_agentic_coding_for_2602_10975,tur_k_ingbench_a_challenge_2403_11905,bearcubs_a_benchmark_for_computer_2503_07919,towards_adaptive_ml_benchmarks_web_2509_09321,from_static_benchmarks_to_dynamic_2602_23729,cloud_opsbench_a_reproducible_benchmark_2603_00468}.
Recent evaluations make this mismatch explicit. For example, 
SWE-bench~\cite{jimenez2024swe}, BigCodeBench~\cite{zhuo2025bigcodebench} and LiveClowBench~\cite{long2026liveclawbench} show that coding capability depends on repository-level context, executable environments, and realistic modification constraints, while MultiChallenge~\cite{deshpande2025multichallenge} shows that dialogue evaluation must capture inferential memory, revision, and consistency across turns.
Together, these limits shift the central question from whether stronger models matter to how model competence is converted into dependable execution.
Agent evaluation must therefore consider task duration, step count, environmental uncertainty, tool-use complexity, state persistence, and recovery demand.
For example, a recent time-horizon study~\cite{kwa2025measuring} evaluates agents by the duration of human tasks they can complete at a fixed success probability, rather than by single-shot accuracy alone.
This framing makes long-horizon reliability central: progress depends not only on model competence, but also on the runtime that turns competence into sustained action, including what the model observes, how context is constructed, which actions are available, where state is preserved, and how errors are detected or repaired.
This helps explain why agent engineering has moved from eliciting isolated model responses toward designing the surrounding execution environment.

\section{Paradigm Shifts in Agent Engineering}
\label{sec:sec4:path}
\begin{figure}[t]
\centering
\includegraphics[width=\linewidth]{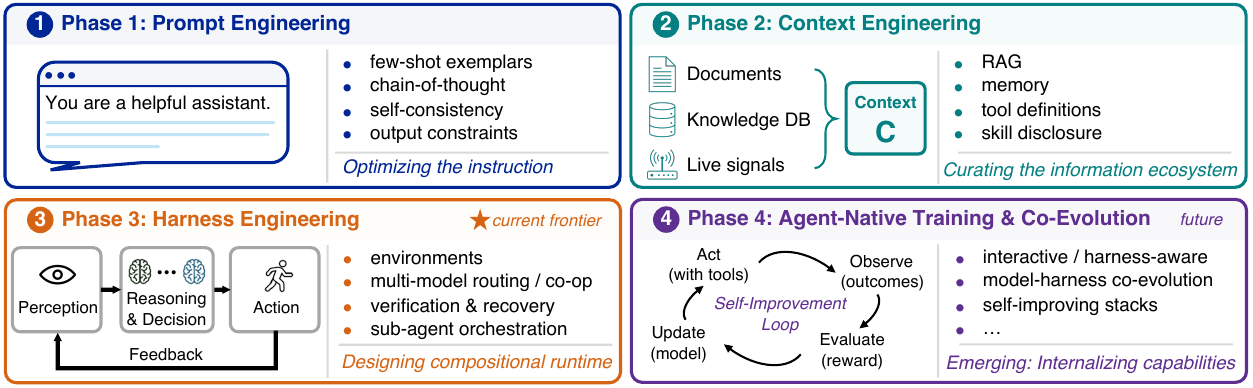}
\caption{\small{Four paradigms of agent engineering. The main locus of effort shifts from eliciting model behavior, to organizing context, stabilizing execution, composing and learning multi-model runtimes, and training or co-evolving agentic behavior.}}
\label{fig:four-paradigms}
\end{figure}

The limits of model-centric scaling in Sec.~\ref{sec:fm} raise a more precise question for agent systems: where is reliable agentic behavior actually produced?
Early LLM-based systems placed much of this burden on prompting, assuming that latent model capabilities could be elicited by effective instructions.
As tasks required external knowledge, tool use, memory, and intermediate artifacts, the focus shifted to agentic workflows and context engineering.
When these workflows became longer, more stateful, and more failure-prone, the bottleneck moved from organizing information for the model to controlling execution around the model, elevating the harness from an implementation detail to a first-class design object.
More recently, verification and recovery have also become training targets, suggesting that some agentic behaviors may be internalized rather than externally scaffolded.
These phases coexist in present systems, but together they reveal a migration of bottlenecks from prompt elicitation, to context and workflow organization, to harness-level execution control, to compositional and learnable multi-model runtimes, and finally toward agent-native training and model--harness co-evolution.
Fig.~\ref{fig:four-paradigms} summarizes this migration as a change in the locus of engineering effort, not as a claim that later paradigms replace earlier ones.

\subsection{Phase 1: Prompt Engineering}
\label{sec:phase1}

Phase 1 treated the prompt as the main interface through which latent model capabilities could be elicited and controlled.
This view was established by in-context learning and then extended through zero-shot and few-shot prompting, chain-of-thought prompting, self-consistency, tree-style reasoning, self-refinement, ReAct-style reasoning-action traces, and automatic prompt optimization~\cite{brown2020gpt3,kojima2022large,wei2022chain,wang2022selfconsistency,yao2023tree,madaan2023self,yao2023react,sahoo2024systematic}.
These methods made prompting a practical mechanism for task specification, reasoning elicitation, output-format control, and behavioral steering.
More recent agent-oriented reasoning work broadens this phase from single-chain elicitation toward structured reasoning and planning. In software tasks, multi-agent optimization and question-driven self-QA extend prompting toward collaborative design reasoning \cite{peng2026beyond}, \cite{liu2026quality}. Self-evolving and graph-structured multi-agent methods further explore reasoning as an adaptive collaboration process rather than a linear trace alone \cite{peng2026sage}, \cite{hao2026brain}. Other work makes reasoning traces operational for failure management \cite{zhang2026efficient}, introduces reflection, branching, and rollback into web-agent reasoning\cite{hu2025webcot}, and uses reasoning gates to decide when web agents should continue or be constrained \cite{kumar2025throttling}. Heterogeneous-model assembly and intent-level skill abstraction also connect prompt-level reasoning with planning and computer-use skill organization \cite{zhu2026symphony}, \cite{lee2026intentcua}, while agentic software-architecture studies frame this progression as a shift from prompt-response interaction to goal-directed systems \cite{alenezi2026prompt}.
They are therefore essential to early agent systems, but their limitation is equally important for this survey's argument.
Prompt engineering primarily addresses an \emph{expression and elicitation} problem: it improves how a task is posed to the model and how the model's existing capabilities are invoked.
It does not reliably provide knowledge absent from the model, maintain dynamically changing task state, validate external actions, or recover from failures over long execution trajectories.
This motivates the next shift, from asking how to phrase the instruction to asking what information environment should surround each model call.

\subsection{Phase 2: Workflows and Context Engineering}
\label{sec:phase2}

Phase 2 shifted the engineering focus from prompt design to agentic workflow orchestration and context management.
This shift addressed two limitations left by prompting: the model may lack task-relevant knowledge, and the information needed during execution may change as the environment responds.
Agentic workflows respond by sequencing model calls, retrieval, tool use, memory access, intermediate artifacts, and branching logic around the model \cite{agashe2025agent,chen2026beyond,chen2026solagent,huang2026tracecoder,chen2026siliconmind}.
Recent systems make this workflow view concrete in software and tool-use settings. SGAgent decomposes repository-level repair into suggestion-guided multi-agent collaboration~\cite{zhang2026sgagent}, while studies of agentic coding-tool configuration show that performance depends not only on the base model, but also on workflow and tool settings~\cite{galster2026configuring}. Tool-use-oriented work further synthesizes tool-use trajectories through multi-agent role-playing~\cite{li2025close} and studies extended tool-integrated reasoning as a way to scale agentic workflows beyond isolated tool calls~\cite{zhang2026aster}.
Within these workflows, context engineering provides the central technical perspective: context is no longer a static prompt string, but a dynamically assembled runtime object.
Formally, instead of assuming $C$=$\mathrm{prompt}$, context is treated as:

\begin{equation}
    C = A(c_1, c_2, \dots, c_n),
\end{equation}
\noindent where $A$ denotes a high-level assembly function that combines contextual components $c_i$, including instructions, retrieved knowledge, tool descriptions and outputs, memory records, task state, intermediate artifacts, and the current query, into the final context $C$.
Under this view~\cite{mei2025survey}, the engineering problem changes from optimizing the wording of a prompt to optimizing the functions that retrieve, select, compress, format, and refresh information during execution:
\begin{equation}
    F^* = \arg\max_F \mathbb{E}_{\tau \sim T}\left[ \mathrm{Reward}\!\left(P_\theta\!\left(Y \mid C_F(\tau)\right),\, Y_\tau^*\right) \right],
\end{equation}
\noindent where $F$ denotes the set of context-construction functions, $C_F(\tau)$ is the context produced for task instance $\tau$, and $Y_\tau^*$ denotes the desired or reference outcome.
The objective is to maximize expected task quality under the constructed context, rather than to optimize a single instruction in isolation.

Agentic workflows therefore reframed the core engineering question from how to write better instructions to how to construct, organize, and update the information and tool-use environment available during execution.
This development can be read through three closely related directions.
The first focused on external information access.
Retrieval-based methods such as RAG~\cite{lewis2020rag,gao2023retrieval} introduced a practical mechanism for exposing non-parametric knowledge to the model, while retrieval-augmented architectures such as Fusion-in-Decoder~\cite{izacard2021leveraging}, RETRO~\cite{borgeaud2022improving}, and Atlas~\cite{izacard2023atlas} further strengthened this paradigm.
Later systems such as RAPTOR~\cite{sarthi2024raptor}, GraphRAG~\cite{edge2024local}, and HippoRAG~\cite{gutierrez2024hipporag} extended retrieval from flat passage lookup to richer pipelines based on hierarchical summarization, graph construction, and relation-aware memory organization.

The second direction focused on systematic context management.
Here the question is not only what to retrieve, but also when to inject information, how to compress it, how to refresh it, and how to preserve task-relevant state over long-horizon execution.
This shift is reflected in methods such as ACON~\cite{kang2025acon}, which formulates context compression as an optimization problem, ARC~\cite{yao2026arc}, which treats context as a dynamically managed internal state updated through reflection, and ContextBudget~\cite{wu2026contextbudget}, which makes compression decisions under explicit context-window constraints.
Related work further examines context maintenance in software and long-horizon settings, including CAT~\cite{liu2025context}, which elevates context maintenance into a callable tool within the agent loop, and Compressing Code Context for LLM-based Issue Resolution~\cite{jia2026compressing}, which studies how to distill and preserve task-relevant code context under limited budgets.

The third direction treated context itself as an explicit object of evaluation and optimization.
Benchmarks such as ContextBench~\cite{li2026contextbench}, SWE Context Bench~\cite{zhu2026swe}, LoCoBench-Agent~\cite{qiu2025locobench}, and AgentLongBench~\cite{fang2026agentlongbench} made context retrieval, retention, and utilization measurable research targets.
More recent work such as ACE~\cite{zhang2025agentic} and MCE~\cite{ye2026meta} further treats contexts, and even context-engineering strategies, as adaptive optimization targets.
Accordingly, Phase 2 can be read as a progression from external information access, to systematic context management, and finally to explicit context evaluation and adaptive context optimization.

Yet even well-engineered workflows and context do not by themselves guarantee reliable agency.
They arrange information, tools, and intermediate steps around the model, but they do not fully specify how the overall process should remain stable, verifiable, and recoverable.
As tasks became increasingly tool-augmented, stateful, and failure-prone, the bottleneck shifted from managing the information and workflow environment to designing the execution environment itself.
Context did not disappear; it became one core component within a broader execution layer that must also manage action, state persistence, and verification.

\subsection{Phase 3: Harness Engineering}
\label{sec:phase3}
This phase begins when the central bottleneck is no longer only how a workflow assembles information and tool calls, but how the agent is controlled across a multi-step execution trajectory.
Agentic workflows improve what enters the context window and which tools are invoked, but many long-horizon failures are not caused by missing information alone.
They are execution failures: the agent loses track of progress, misuses tools, drifts from the original objective, repeats unproductive steps, or fails to recover after an error.

Harness engineering emerges from this execution-level bottleneck.
As defined in Sec.~\ref{sec:sec2.4-harness}, the harness is the structured runtime layer that organizes and stabilizes agent execution.
It determines what the agent observes, what actions it may take, what state is carried forward, how control advances, and how failures are detected, constrained, or repaired.
Workflow and context design remain important, but they become components within a broader execution layer that also manages observation, action, persistent state, verification, and governance.

\noindent\textbf{Evidence that harness design matters.}
The case for harness design is no longer based only on engineering intuition or isolated examples.
SWE-agent~\cite{yang2024sweagent} showed that redesigning the agent-computer interface alone can substantially improve coding-agent performance under a fixed model.
NLAH framed harness modules as portable and inspectable artifacts, and reported controlled ablations indicating that their contributions are measurable and additive~\cite{pan2026nlah}.
Meta-Harness went one step further by treating harness optimization itself as a search problem, showing that automatically improved harnesses can outperform hand-designed baselines on Terminal-Bench~\cite{lee2026metaharness,merrill2026terminalbench}.
Recent works extend this evidence from coding-agent interfaces to runtime orchestration and formal control mechanisms~\cite{shen2026dova,utility_guided_agent_orchestration_for_2603_19896,sulc2026differentiable}. Other systems treat memory, protocol interoperability, contextual problem enhancement, and enterprise context lifecycles as harness-level design objects~\cite{memma_coordinating_the_memory_cycle_2603_18718,structurally_aligned_subtask_level_memory_2602_21611,mcp_vs_rag_vs_nlweb_2511_23281,codescout_contextual_problem_statement_enhancement_2603_05744,ldp_an_identity_aware_protocol_2603_08852,context_engineering_from_prompts_to_2603_09619}.
These results suggest that harness design has become a first-class optimization surface rather than a secondary implementation detail. More experimental results can be found in Sec.~\ref{sec:sec7:eval}.

\noindent\textbf{Industrial and ecosystem perspectives.}
The same transition is visible in industrial systems.
Anthropic's public guidance emphasizes minimal, legible tools and disciplined runtime behavior~\cite{claudecode2025,anthropic2024building,anthropic2025harness}.
OpenAI's guidance emphasizes environment design, structured artifacts, and reusable agent infrastructure~\cite{openai2026harness,openai2025agentguide,openai2025agentssdk}.
Microsoft's Magentic-One highlights multi-agent orchestration for complex web and file tasks~\cite{fourney2024magenticone}, while open-source systems~\cite{wang2024openhands,opensquilla2026}, \eg, OpenHands, expose harness itself as inspectable code.

At the ecosystem level, recent protocol-centered benchmarks reinforce the same shift by evaluating whether agents can invoke real services under realistic tool-routing conditions.
MCPWorld~\cite{yan2025mcpworld}, MCP-Atlas~\cite{bandi2026mcpatlas}, MCPAgentBench~\cite{liu2025mcpagentbench}, and OSWorld-MCP~\cite{jia2025osworldmcp} move the discussion from abstract protocol design to measurable runtime behavior.
Together, these systems and benchmarks suggest that harness engineering is becoming not only an engineering practice, but also a shared layer of infrastructure, evaluation, and design philosophy.

\noindent\textbf{Design principles.}
Across papers and systems, several high-level principles recur:

\begin{itemize}[leftmargin=*]
  \item \textbf{Legibility:} the runtime should expose the right state at the right level of abstraction.
  \item \textbf{Mechanical enforcement:} constraints that matter for safety, correctness, or reproducibility should be enforced by the runtime when possible, rather than delegated entirely to prompt obedience.
  \item \textbf{Verification in the loop:} long-horizon autonomy without intermediate checks is structurally brittle.
  \item \textbf{Explicit artifacts:} plans, logs, diffs, summaries, and other intermediate products should exist as inspectable objects that can be reused, audited, or handed off.
\end{itemize}

These principles make the harness concrete rather than metaphorical.
It must expose observations, assemble context, organize control, mediate actions, persist state, and enforce verification and governance as a coupled runtime system.
Sec.~\ref{sec:sec5:anatomy} therefore turns the phase-level argument into an anatomy of the six harness components.

\noindent\textbf{Multi-model harnesses.}
Many recent systems no longer treat one model as the sole cognitive engine for every step.
Instead, the harness composes heterogeneous models for planning, coding, tool use, verification, retrieval, and domain-specific subtasks~\cite{fourney2024magenticone,zhu2026symphony,openai2025agentssdk,wang2024openhands,agashe2025agent}.
This changes the control loop from ``one model iterates until done'' to ``the runtime decides which model acts next, with what context, and under what constraints.''
Representative patterns include planner--executor--verifier decomposition, specialist routing, debate or committee-style validation, and handoffs among sub-agents~\cite{fourney2024magenticone,zhang2026sgagent,li2025close,vallabhaneni2026ai}.
From the harness-anatomy view in Sec.~\ref{sec:sec5:anatomy}, multi-model design~\cite{mcp2025,google2025a2a,yan2025mcpworld,jia2025osworldmcp}. primarily stresses the control loop $\mathcal{L}$, but it also reshapes the context manager $\mathcal{C}$, action interface $\mathcal{I}_{\mathrm{act}}$, and verification layer $\mathcal{V}$ because different models may observe, act, and judge under different scopes and permissions.

\noindent\textbf{Learnable harnesses.}
In parallel, the harness itself is becoming an optimizable object.
NLAH~\cite{pan2026nlah} treats harness logic as editable and portable runtime artifacts; Meta-Harness~\cite{lee2026metaharness} searches over harness configurations; and Agentic Harness Engineering (AHE)~\cite{lin2026agenticHarnessEngineering} evolves harness components from observability-driven feedback while holding the base model fixed.
These systems differ in mechanism---manual editing, search, or trace-driven adaptation---but they share one implication: runtime policies for routing, tool exposure, memory use, and verification can be improved as directly as prompts once were.

Together, multi-model composition and learnable runtime policies mark a qualitative shift in what counts as an agent system.
A lightweight prompt-driven loop can still behave like an agent on short horizons, but dependable long-horizon task completion increasingly requires designing a compositional runtime over multiple models, with explicit orchestration, verification, and adaptation policies.

\subsection{Phase 4: Agent-Native Training and Co-Evolution}
\label{sec:phase4}

Phase~4 begins once the harness is viewed not only as a hand-stabilized runtime, but as a compositional and increasingly learnable system over one or more models (Sec.~\ref{sec:phase3}).
The central question is therefore twofold: which agentic behaviors should be \emph{internalized} into model parameters, and how should the model and harness \emph{co-evolve} over deployment without sacrificing safety or inspectability.

\noindent\textbf{Internalization through Interactive Training.}
The first direction internalizes agentic behavior into the model itself.
Rather than relying solely on prompts, workflows, or runtime orchestration, models are increasingly trained to plan, use tools, verify intermediate states, and recover from errors in interactive environments~\cite{webagent_r1_training_web_agents_2505_16421,workforceagent_r1_incentivizing_reasoning_capability_2505_22942,lai2025computerrl,tgpo_tree_guided_preference_optimization_2509_14172,zhou2025esearch,ding2026dynaweb}.

Recent work reflects two closely related tendencies.
The first strengthens reasoning-to-action behavior through reinforcement learning.
Examples include DeepSeekMath~\cite{shao2024deepseekmath}, DeepSeek-R1~\cite{deepseek2025r1}, and DAPO~\cite{yu2026dapo}, which treat multi-step reasoning, action selection, and verification as trainable behaviors rather than purely prompt-induced ones.
The second reduces train-test mismatch by training agents in environments closer to deployment.
Examples include ProRL~\cite{zhang2026prorl}, WebRL~\cite{qi2024webrl}, ComputerRL~\cite{lai2025computerrl}, Environment Tuning~\cite{lu2025don}, daVinci-Dev~\cite{zeng2026davinci}, and Kimi-Dev~\cite{yang2025kimi}.
Together, these lines suggest that behaviors first implemented externally---planning, tool invocation, reflection, and recovery---may gradually become partially learned inside the model.
Internalization shifts the division of labor rather than removing the harness: more short-horizon behavior may move into model parameters, while the runtime still supplies environment access, state, and safety control.

\noindent\textbf{Co-Evolution and Self-Improvement.}
The second direction extends agent engineering from one-time training to ongoing improvement of the full stack.
Here the model, harness, and update policy may all change over deployment, using execution feedback to decide which changes to keep, revise, or roll back~\cite{wu2025evolver,zhai2025agentevolver,zhang2025darwin,lin2026agenticHarnessEngineering,karten2026continual}.
The goal is not only to move behavior into parameters, but to improve \emph{how} the combined system learns from experience.

Several recent lines make this distinction concrete.
Experience-driven systems such as EvolveR~\cite{wu2025evolver} and AgentEvolver~\cite{zhai2025agentevolver} treat interaction trajectories as reusable learning signals through self-questioning, navigation, and attribution.
Continual Harness~\cite{karten2026continual} and reward-free self-evolution~\cite{zhang2026training} explore online adaptation without relying on dense external rewards at inference time.
Harness-side adaptation, exemplified by AHE~\cite{lin2026agenticHarnessEngineering}, shows that runtime components can evolve even when the base model remains fixed.
More ambitiously, recursive self-improvement systems such as SICA~\cite{robeyns2025self}, Darwin G\"odel Machine~\cite{zhang2025darwin}, and Hyperagents~\cite{zhang2026hyperagents} suggest that the improvement mechanism itself may become modifiable over time.

We separate three layers that are often conflated under ``self-evolve''.
\emph{Multi-model harnesses} define \emph{who} performs each runtime role.
\emph{Learnable harnesses} define \emph{how} runtime policies are optimized.
\emph{Co-evolution} defines \emph{when and how} the model, harness, and improvement loop are jointly updated from deployment experience.
These layers are complementary rather than interchangeable: compositional runtimes create the structure in which specialization and delegation become possible; learnable harnesses make runtime adaptation explicit; co-evolution governs long-horizon improvement under verification, safety and cost constraints.

\section{Anatomy of the Execution Harness}

\label{sec:sec5:anatomy}

Harness engineering shifts the optimization target from isolated prompts or workflows to the runtime that stabilizes agent execution.
Following the formalization in Sec.~\ref{sec:sec2.4-harness}, this runtime can be decomposed into six components:
$\mathcal{H} = \langle \mathcal{I}_{\mathrm{obs}}, \mathcal{C}, \mathcal{L}, \mathcal{I}_{\mathrm{act}}, \mathcal{S}, \mathcal{V} \rangle$.
The decomposition is not intended as a software package diagram.
Rather, it identifies the runtime responsibilities that repeatedly determine whether model capability becomes reliable task completion: what the model observes, what enters context, how execution advances, which actions are available, what state persists, and how the run is checked or constrained.

\subsection{Observation Interface}
\label{sec:anatomy-observation-interface}

The observation interface $\mathcal{I}_{\mathrm{obs}}$ determines which environment signals are exposed to the model and how those signals are rendered.
It converts external state, such as terminal output, file diffs, screenshots, web DOMs, API responses, event streams, and logs, into observations that can be consumed by the current model call.
Its design space includes three recurring questions: which state is relevant, at what abstraction level it should be represented, and when the observation should be refreshed.
These choices matter because many long-horizon failures are not failures of reasoning alone.
They are also failures of legibility: the needed state is absent, buried in noise, stale, or represented at a level that does not support the next decision.

Representative systems make this point concrete.
SWE-agent showed that redesigning the agent-computer interface can substantially improve coding-agent performance under a fixed base model~\cite{yang2024sweagent}.
In web and desktop settings, benchmarks such as WebArena and OSWorld likewise reveal that success depends on whether visually and structurally complex interface state is converted into a form the model can actually use~\cite{zhou2024webarena,xie2025osworld}.
The general principle is therefore not to expose all available state, but to expose decision-relevant state in a faithful and usable representation.
The dominant trade-off is richness versus tractability: rich observations improve grounding, but increase context cost and distractors; compressed observations are easier to process, but may discard task-critical evidence.
An open problem is to design observation interfaces that remain faithful and decision-useful under partial observability, multimodal state, and long trajectories.

\subsection{Context Manager}
\label{sec:anatomy-context-manager}

Context management first emerged as a central concern in agentic workflows.
Within the harness, it becomes one runtime component among observation, control, action, persistence, and verification.
The context manager $\mathcal{C}$ determines which available information enters the current model call and in what form.
It selects, compresses, orders, and refreshes observations, tool outputs, retrieved evidence, memory records, summaries, instructions, and task artifacts before they become the working context for the next step~\cite{codescout_contextual_problem_statement_enhancement_2603_05744,profile_then_reason_bounded_semantic_2604_04131,mem2actbench_a_benchmark_for_evaluating_2601_19935,memory_for_autonomous_llm_agents_2603_07670}.
Its main design choices concern inclusion, representation, refresh policy, and the amount of shared state exposed to active agents or sub-agents.
As context engineering suggests, long-horizon performance depends less on simply increasing prompt length than on maintaining coherent task state over time~\cite{anthropic2025context,context_engineering_from_prompts_to_2603_09619,explore_with_long_term_memory_2601_10744,memorycd_benchmarking_long_context_user_2603_25973,hippocamp_benchmarking_contextual_agents_on_2604_01221}.

Several implementation patterns recur.
Retrieval-based systems bring in external documents or stored state on demand.
Memory-oriented systems such as MemGPT separate the active context from a larger external memory~\cite{packer2023memgpt}.
Industrial harnesses increasingly externalize task state into explicit artifacts and selectively resurface those artifacts, rather than relying on a single ever-growing dialogue trace~\cite{openai2025agentguide}.
The dominant trade-off is fidelity versus manageability: raw histories preserve detail but scale poorly, whereas summaries and retrieved context are cheaper but can omit or distort important state.
Thus, the crucial distinction is not between long and short prompts, but between monolithic and managed context.
A key open problem is how to preserve summary faithfulness and state integrity while keeping context cost bounded, especially when context repair must interact with verification and recovery in long-horizon settings~\cite{maharana2024locomo}.

\subsection{Control Loop}
\label{sec:anatomy-control-loop}
The control loop $\mathcal{L}$ organizes execution across steps, tools, and possible handoffs.
It turns observation, reasoning, action, and feedback into a runnable process.
This component determines whether the agent follows a simple perceive-act cycle, a ReAct-style loop, a plan-execute-verify routine, or a hierarchical and multi-agent workflow~\cite{shen2026dova,utility_guided_agent_orchestration_for_2603_19896,sulc2026differentiable,zhu2026symphony,agashe2025agent}.
The main design questions are how control is divided between the model and the runtime, when plans are created or revised, when delegation is introduced, whether coordination is sequential or parallel, and when execution should stop.
Long-horizon success depends not only on the model's reasoning quality, but also on whether the runtime keeps execution stable under uncertainty.

Existing systems occupy different points in this space.
Some retain lightweight iterative loops, while others impose explicit planner-executor-verifier decomposition or multi-agent coordination.
Recent harness-oriented work makes orchestration itself an optimization target: NLAH treats harness logic as an editable artifact, and Meta-Harness treats harness configuration as a searchable design space~\cite{pan2026nlah,lee2026metaharness}.
This layer is therefore not merely about adding steps.
It is about choosing how much structure to impose on the trajectory.
The dominant trade-off is adaptability versus stability: freer loops can respond to unexpected states, but are more prone to drift, repeated failure, and coordination overhead; stronger orchestration improves reliability, but can reduce efficiency or overconstrain exploration.
A standing challenge is to design control policies that remain robust across horizons and domains without making delegation, verification, and recovery prohibitively expensive.

\subsection{Action Interface}
\label{sec:anatomy-action-interface}
The action interface $\mathcal{I}_{\mathrm{act}}$ maps model outputs to executable operations.
It defines what the agent can do, how actions are specified, which permissions apply, and how action results are returned as subsequent observations.
Recent tool-use studies further show that action-interface quality is a major source of agent reliability. Diagnostic work on tool invocation failures identifies cases where agents fail not because a tool is absent, but because the tool is poorly selected, invoked, or integrated into the execution trajectory~\cite{huang2026agents}. ET-Agent studies behavior calibration for tool-integrated reasoning~\cite{chen2026agent},  and ToolTok represents tools as tokens to improve efficiency and generalization in GUI agents~\cite{wang2026tooltok}.
From a harness perspective, this layer shapes the agent's effective action space.
Its design space spans tool granularity (low-level environments versus high-level APIs), tool specification (free-form commands versus structured schemas), routing and interoperability (local tools versus protocol-based ecosystems such as MCP), and governance (permissions, side-effect control, and invocation constraints).
Existing work shows that performance depends not only on whether tools exist, but on how the action interface makes them usable, composable, and governable.

Representative implementations range from terminals and browsers to structured callable APIs.
SWE-agent is a canonical example of observation-action co-design: redesigning the agent-computer interface changes both what the model sees and how it acts, producing gains under a fixed base model~\cite{yang2024sweagent}.
Protocol-oriented infrastructures such as MCP move tool access toward a standardized interface layer across heterogeneous services~\cite{mcp2025}, while benchmarks such as MCPWorld and OSWorld-MCP test whether agents can invoke such services reliably in realistic environments~\cite{yan2025mcpworld,jia2025osworldmcp}.
The dominant trade-off is flexibility versus controllability: low-level tools are general but difficult to use robustly, whereas high-level tools improve reliability but may narrow behavior too aggressively.
An open question is how to design action abstractions that remain expressive and governable when tool use is coupled with verification, sandboxing, and recovery over long horizons.

\subsection{State and Artifact Store}
\label{sec:anatomy-state-artifact-store}

The state and artifact store $\mathcal{S}$ persists execution state across steps, sessions, and subtasks.
It provides continuity beyond the active context window by storing task progress, traces, plans, checkpoints, diffs, generated files, memory records, and other reusable artifacts.
Its design space includes the granularity of stored state, the scope of persistence, the storage form, and the update policy.
Long-horizon agents often fail not because no state is stored, but because the wrong state is preserved, the right state cannot be retrieved, or stale state is treated as current.

Several strategies recur in the literature.
Some systems rely on session-level histories and checkpoint stores.
Memory-oriented approaches, such as MemGPT, introduce explicit long-term memory beyond the current context~\cite{packer2023memgpt}.
Artifact-centered harnesses track state through logs, diffs, checkpoints, and inspectable runtime objects~\cite{openai2025agentguide}.
What these approaches share is a move from transient interaction traces toward reusable system state.
The dominant trade-off is completeness versus usability: richer state improves continuity, auditability, and handoff, but increases retrieval burden, noise, and the risk of stale memory.
The practical challenge is not to store more, but to decide what deserves persistence, what should be compressed, and what should be discarded.
An important open problem is how to maintain state fidelity while supporting rollback, delegation, and memory reuse without accumulating drift or obsolete information~\cite{maharana2024locomo}.

\subsection{Verification and Governance}
\label{sec:anatomy-verification-governance}
The verification and governance layer $\mathcal{V}$ checks, constrains, and repairs execution during runtime.
Verification includes tests, assertions, verifier models, judge signals, and other mechanisms for estimating whether execution is progressing correctly.
Governance includes approval gates, sandboxing, budget control, rollback, retry, escalation, and safe termination.
This view is reflected in recent multi-agent governance works. For example, AI Committee uses multiple agents for validation and remediation of web-sourced data \cite{vallabhaneni2026ai}, while act-or-refuse learning studies when agents should proceed, abstain, or stop during safe multi-step tool use \cite{agarwal2026learning}. These examples show that governance is not only a deployment constraint, but also an explicit decision layer within agent execution.
These two roles are tightly coupled: verification produces evidence about the run, while governance determines what the harness is allowed or required to do with that evidence.
Reliable agents therefore depend not only on strong reasoning and rich tools, but also on whether checks and constraints are mechanically enforced rather than left entirely to prompt obedience.

Representative systems make this dual role clear.
In coding agents, tests, linters, and assertions provide relatively strong verification signals, while rollback and sandboxed execution contain side effects~\cite{jimenez2024swe,merrill2026terminalbench}.
In web, desktop, and other open-ended environments, governance becomes more central because actions may be partially irreversible and clean oracles are often unavailable~\cite{zhou2024webarena,xie2025osworld}.
The dominant trade-off is autonomy versus robustness: looser constraints permit broader exploration, but increase the risk of harmful actions and unrecoverable drift; tighter governance improves safety and recoverability, but can slow execution or block useful behavior.
An open problem is to design verification and governance mechanisms that are selective and cost-aware, so the harness can distinguish recoverable local errors from deeper task-level collapse without over-triggering interruption or rollback.

\subsection{Cross-Layer Interactions in the Harness}
\label{sec:anatomy-cross-layer-interactions}

Although the six components are analytically separable, they do not operate independently.
Design choices in one component often reshape the burden on others.
The observation interface $\mathcal{I}_{\mathrm{obs}}$ and context manager $\mathcal{C}$ are tightly coupled: richer observations can improve grounding, but they also increase the cost of selection, compression, and formatting before information enters the active context~\cite{kang2025acon}.
The action interface $\mathcal{I}_{\mathrm{act}}$ interacts directly with verification and governance $\mathcal{V}$: more expressive actions expand capability, but require stronger permission control, sandboxing, rollback, and auditing.
The state and artifact store $\mathcal{S}$ feeds back into context and verification because persistent plans, logs, checkpoints, and artifacts determine both what can be resurfaced to the model and what evidence is available for judging progress~\cite{packer2023memgpt,maharana2024locomo}.

Harness design is therefore a coupled systems problem rather than the independent optimization of six modules.
Improving one layer can shift risk elsewhere: stronger compression can reduce cost while weakening downstream verification; richer actions can improve task coverage while increasing governance pressure; more persistent state can improve continuity while also introducing stale or conflicting evidence.
This coupling makes task structure part of harness design itself.
Different domains, horizons, oracle strengths, and autonomy requirements place different pressure profiles over $\mathcal{I}_{\mathrm{obs}}$, $\mathcal{C}$, $\mathcal{L}$, $\mathcal{I}_{\mathrm{act}}$, $\mathcal{S}$, and $\mathcal{V}$.
The same anatomy therefore becomes a way to read the task landscape: tasks differ by which runtime responsibilities they stress and which configuration choices become decisive.

\section{Task Landscape and Harness Configuration}
\label{sec:tasks}

Task structure determines which parts of the execution harness become performance-critical.
Seen through the harness anatomy, a task is not merely an application label but a pressure profile over observation, context, control, action, state, and governance.
Long horizons, partial observability, weak feedback, irreversible actions, and autonomy requirements shift pressure toward different configuration choices, including context selections, action abstractions, verifier loops, checkpoints, permission gates, and escalation rules.
The central question is therefore which runtime responsibility becomes the limiting factor under a given task condition.

\subsection{A Harness-Aware Task Taxonomy}
\label{sec:tasks:framework}

Agent tasks differ not only by application label, but by structural properties that determine which harness components limit performance, reliability, cost, or safety.
We use three dimensions to characterize these pressures: \textbf{task horizon, environment type, and autonomy level.}
Together, they identify the primary bottleneck: the component or small set of components where failures most often concentrate.

\paratitle{Complexity and task horizon.}
Tab.~\ref{tab:task-complexity} summarizes four coarse levels.
The most consequential transition is usually from L2 to L3.
Single-step and short multi-step tasks can often be handled with local reasoning, lightweight action access, and local checks.
Long-horizon tasks create sustained pressure on the Context Manager, State and Artifact Store, and Control Loop because plans, intermediate artifacts, failed attempts, and partial results must survive beyond one prompt window.
At L4, open-ended monitoring or exploration additionally requires budget control, stopping criteria, and escalation policies, shifting the bottleneck toward explicit Verification and Governance rather than generation quality alone.

\begin{table}[t]
\centering
\caption{\small{Harness-aware task complexity levels and their primary bottlenecks.}}
\label{tab:task-complexity}
\begin{tabular}{>{\centering\arraybackslash}m{0.6cm}
>{\centering\arraybackslash}m{1.7cm}
>{\centering\arraybackslash}m{2.2cm}
>{\centering\arraybackslash}m{2.5cm}}
\toprule
\textbf{Level} & \textbf{Description} & \textbf{Examples} & \textbf{Bottleneck} \\
\midrule
\rowcolor{gray!15}
L1 & Single-step & Search, translate, calculate & Context Mgr.; Verif. \& Gov. \\
L2 & Multi-step & Form filling, code generation & Act. Interface; State Store; Verif. \& Gov. \\
\rowcolor{gray!15}
L3 & Long-horizon & Repo-scale coding, research & Context Mgr.; State Store; Ctrl. Loop \\
L4 & Open-ended & Monitoring, auto exploration & Verif. \& Gov.; Ctrl. Loop \\
\bottomrule
\end{tabular}
\end{table}

\paratitle{Environment type.}
Environment type determines what the harness must observe and which actions it can expose~\cite{finmcp_bench_benchmarking_llm_agents_2603_24943,agentdrive_an_open_benchmark_dataset_2601_16964,cap_x_a_framework_for_2603_22435,prosoftarena_benchmarking_hierarchical_capabilities_of_2601_02399,abc_bench_benchmarking_agentic_backend_2601_11077,enterpriseops_gym_environments_and_evaluations_2603_13594,phmforge_a_scenario_driven_agentic_2604_01532}.
Terminals and repositories stress the Action Interface, Observation Interface, and Verification and Governance components because commands, diffs, logs, and tests can often be wrapped as structured actions, observations, and verifier signals.
Browser and desktop environments add visual or DOM grounding, session state, and persistent side effects, increasing pressure on observation construction, action abstraction, and permission boundaries.
Knowledge environments, including web search, literature retrieval, and structured databases, shift the bottleneck toward the Context Manager and State and Artifact Store because the main challenge is managing evidence quality, provenance, and synthesis.
Physical environments introduce real-time constraints and irreversibility, making the Control Loop and Verification and Governance more central than in purely digital settings.
Social environments add norms, negotiation, and strategic responses, which raises the value of richer observation design and conservative escalation.

\paratitle{Autonomy level.}
Autonomy cuts across domains.
Human-in-the-loop settings can route high-risk decisions through approval gates.
Semi-autonomous systems delegate routine actions but escalate when uncertainty rises.
Fully autonomous systems must absorb more of that burden inside the harness through verification, rollback, logging, and fail-safe termination.
Autonomy is best understood as a multiplier on harness requirements: the more independently the system is expected to act, the more the Observation Interface and Verification and Governance move from optional guardrails to primary bottlenecks.

Taken together, these dimensions define a pressure profile over the six harness components.
Task horizon mostly stresses context, state, and control; environment type mostly stresses observation, action, and safety boundaries; autonomy mostly stresses verification, logging, and recovery.
This view turns task taxonomy into harness configuration: the key question is not simply which application domain a task belongs to, but which runtime responsibilities become bottlenecks under its structural pressures.
The next subsection uses representative domains as case studies to illustrate this bottleneck migration in concrete settings.

\subsection{Harness Adaptation by Domain}
\label{sec:tasks:domains}

The following domains should be read as \emph{instantiations} of the pressure-profile view above, not as a separate domain-only taxonomy.
Each domain combines horizon, environment, oracle strength, irreversibility, and autonomy in a different way, thereby shifting the primary harness bottleneck.
Software engineering is verification-dominant; web and GUI interaction is grounding-dominant; scientific discovery is synthesis-dominant; medical assistance is safety-dominant; and embodied agents are control-dominant.
These cases show how the same harness anatomy leads to different configuration priorities once task pressures change.
Tab.~\ref{tab:task-harness-mapping} later abstracts these examples into reusable configuration rules.

\paratitle{Software engineering (verification-dominant).}
Software engineering places the primary bottleneck on Verification and Governance~\cite{swe_replay_efficient_test_time_2601_22129,agentstepper_interactive_debugging_of_software_2602_06593,wink_recovering_from_misbehaviors_in_2602_17037,xai_for_coding_agent_failures_2603_05941}.
Builds, unit tests, linters, and execution logs provide mechanical feedback signals that most other domains lack~\cite{jimenez2024swe}.
These strong oracles enable closed-loop generate-test-repair cycles, where verifier output becomes the next iteration's evidence.
Systems such as Claude Code, SWE-agent, and OpenHands further show that redesigning the Action Interface, including file and command interfaces, diff views, and patch inspection, can yield measurable gains at fixed model capability~\cite{claudecode2025,yang2024sweagent,wang2024openhands}.
The State and Artifact Store is equally consequential: expressive action access must be paired with checkpoints, patch artifacts, safe rollback, and bounded side effects.
As tasks scale from snippet-level generation (L2) to repo-scale issue resolution (L3-L4), the Context Manager and State and Artifact Store join the bottleneck set for tracking plans, failed hypotheses, and intermediate artifacts across long trajectories~\cite{menvagent_scalable_polyglot_environment_construction_2601_22859,agentictyper_automated_typing_of_legacy_2602_21251,swe_protege_learning_to_selectively_2602_22124,from_language_to_action_can_2603_03148}.

\begin{itemize}[leftmargin=*]
    \item \textbf{Configuration implication.} Verification-dominant settings turn runtime quality into a closed-loop optimization problem; the harness response is verifier loops, generate-test-repair cycles, and reversible execution.
\end{itemize}

\paratitle{Web and GUI interaction (grounding-dominant).}
Web and GUI agents shift the primary bottleneck from verification to grounding: the core difficulty is constructing observations the model can use and actions it can execute safely~\cite{zhou2024webarena,koh2024visualwebarena,xie2025osworld,jia2025osworldmcp,coact_1_computer_using_multi_2508_03923,the_dawn_of_gui_agent_2411_10323,agenttrek_agent_trajectory_synthesis_via_2412_09605,efficient_agent_training_for_computer_2505_13909}.
The Observation Interface must render screenshots, DOM state, interaction history, and session information into usable signals.
The Action Interface must decide whether actions are exposed as brittle low-level selectors or as structured browser and desktop operations.
The Context Manager then selects and compresses these signals for the current step.
Because many web goals lack a single mechanical oracle, verification is structurally weaker than in coding.
Navigation mistakes, form submissions, and account actions can produce persistent side effects, so Verification and Governance remains part of the bottleneck rather than a secondary concern.
As tasks move from short form filling (L2) through multi-page workflows (L3) to long-lived monitoring (L4), the grounding burden compounds.

\begin{itemize}[leftmargin=*]
    \item \textbf{Configuration implication.} Grounding-dominant settings require co-design of observation and action interfaces; performance hinges on whether the harness exposes the right state in a form that also supports safe, robust action.
\end{itemize}

\paratitle{Scientific discovery and research (synthesis-dominant).}
Scientific agents shift the primary bottleneck to synthesis: the central runtime problem is trustworthy integration of evidence over long horizons~\cite{bran2023chemcrow,roohani2024biodiscovery,trace_a_multi_agent_system_2603_21152,agent_driven_corpus_linguistics_a_2604_07189,scivisagentbench_a_benchmark_for_evaluating_2603_29139,evoscientist_towards_multi_agent_evolving_2603_08127}.
The most stressed components are the Context Manager, State and Artifact Store, and Verification and Governance.
Verification serves a different function here than in coding: rather than closing a test-based repair loop, it must assess provenance, source quality, and reasoning coherence.
Tool-rich systems such as ChemCrow and BioDiscoveryAgent show that the Action Interface can expand capability, but without provenance tracking long-horizon reasoning can degrade into plausible but unsupported narrative.
Research tasks are predominantly L3-L4: literature surveys, hypothesis generation, and experimental design require the harness to externalize evidence, intermediate claims, and artifacts more aggressively than in coding or web settings.

\begin{itemize}[leftmargin=*]
    \item \textbf{Configuration implication.} Synthesis-dominant settings lack closed-loop verification; provenance tracking, intermediate review stages, and artifact-centered memory must substitute for end-state oracles.
\end{itemize}

\paratitle{Medical applications (safety-dominant).}
Medical agents inherit the synthesis demands of research settings but add a qualitatively different constraint: the primary bottleneck shifts to safety~\cite{improving_clinical_diagnosis_with_counterfactual_2603_27820,ccd_cbt_multi_agent_therapeutic_2604_06551,eft_cot_a_multi_agent_2601_17842,medbrowsecomp_benchmarking_medical_deep_research_2505_14963}.
Verification and Governance moves to the top of the harness stack, with the Observation Interface and Context Manager close behind~\cite{li2024agenthospital}.
Patient history, guidelines, and recent findings must be surfaced accurately; consequential actions must be permission-gated; and uncertainty must trigger conservative escalation rather than confident continuation.
The objective is not maximal autonomy, but bounded and inspectable assistance.
In this regime, Verification and Governance is a performance-critical harness component rather than a compliance add-on.

\begin{itemize}[leftmargin=*]
    \item \textbf{Configuration implication.} Safety-dominant settings optimize controlled delegation: approval gates, auditability, and conservative recovery are core harness components, not external overhead.
\end{itemize}

\paratitle{Embodied settings (control-dominant).}
Embodied agents shift the primary bottleneck to real-time control, elevating the Control Loop above other harness components~\cite{wang2023voyager,can_a_robot_walk_the_2603_21723,robosafe_safeguarding_embodied_agents_via_2512_21220,when_should_a_robot_think_2603_16673,agriworld_a_world_tools_protocol_2602_15325}.
High-level language reasoning is too slow for continuous interaction, so the harness typically becomes layered: deliberative planning at the top, reactive control below, and persistent skill or state representations connecting the two.
The State and Artifact Store maintains goals, subgoals, maps, or reusable behaviors.
Verification and Governance is critical because physical actions are often irreversible.
The Action Interface exposes actuators, simulators, or perception modules rather than conventional software APIs.
Embodied tasks span L3-L4 almost exclusively, making long-horizon state externalization a baseline requirement.

\begin{itemize}[leftmargin=*]
    \item \textbf{Configuration implication.} Control-dominant settings push part of the stack below the language loop, motivating tighter integration between harness design, lower-level controllers, and training-time adaptation.
\end{itemize}

\paratitle{Cross-domain synthesis.}
The five domains trace a single analytic thread: the primary bottleneck migrates across harness components as domain constraints change.
It moves from Verification and Governance in coding, through Observation Interface and Action Interface in web/GUI, to Context Manager and State and Artifact Store in research, Verification and Governance in medicine, and Control Loop in embodied settings.
This migration pattern shows that domain labels alone are insufficient descriptors.
What matters for harness configuration is which component absorbs the failure budget.

\subsection{From Task Properties to Harness Configurations}
\label{sec:tasks:patterns}

The domain case studies above illustrate bottleneck migration, but the reusable lesson lies below the domain level.
Across domains, similar task properties induce similar harness responses: long horizons require externalized state, partial observability requires structured observation, strong oracles enable verifier loops, weak or delayed oracles require provenance and review, irreversible actions require governance, and high autonomy requires logging, budgets, and recovery.
Tab.~\ref{tab:task-harness-mapping} summarizes these domain-independent configuration rules.

\begin{table*}[t]
\centering
\caption{\small{Mapping task properties to harness failure pressures and configuration responses.}}
\label{tab:task-harness-mapping}
\small
\setlength{\aboverulesep}{0pt}
\setlength{\belowrulesep}{0pt}
\setlength{\tabcolsep}{1.5pt}
\renewcommand{\arraystretch}{1.08}
\begin{tabular}{cccc}
\toprule
\addlinespace[2pt]
\textbf{Task property} 
& \textbf{Failure pressure}
& \textbf{Harness response} 
& \textbf{Critical components} \\
\midrule
\rowcolor{gray!15}
Long horizon 
& State drift
& Checkpoints, summaries, artifacts 
& $\mathcal{C}$, $\mathcal{S}$, $\mathcal{L}$ \\
Partial observability
& Indirect state 
& Structured observations, grounding, abstraction 
& $\mathcal{I}_{\mathrm{obs}}$, $\mathcal{C}$, $\mathcal{I}_{\mathrm{act}}$ \\
\rowcolor{gray!15}
Strong oracle 
& Checkable outcomes 
& Verifier loops, repair cycles 
& $\mathcal{V}$, $\mathcal{L}$ \\
Weak or delayed oracle 
& Uncertain success 
& Provenance tracking, review, approval 
& $\mathcal{V}$, $\mathcal{C}$, $\mathcal{S}$ \\
\rowcolor{gray!15}
Irreversible actions 
& Persistent side effects 
& Sandbox, gates, rollback 
& $\mathcal{V}$, $\mathcal{I}_{\mathrm{act}}$ \\
High autonomy or low latency 
& Limited human correction 
& Logging, budgets, controllers 
& $\mathcal{V}$, $\mathcal{L}$, $\mathcal{I}_{\mathrm{obs}}$ \\
\bottomrule
\end{tabular}
\end{table*}

\paratitle{Verifier strength determines where configuration effort concentrates.}
Where strong automatic oracles exist, as in verification-dominant software engineering, the harness can invest heavily in closed-loop optimization through verifier loops and repair cycles.
Where oracles are weak or delayed, as in synthesis-dominant research and safety-dominant medicine, the bottleneck migrates upstream toward provenance management, intermediate review, and conservative stopping criteria.
Domains should therefore not be compared only by task success rates; they should also be compared by the quality and latency of feedback signals available to the harness.

\paratitle{Irreversibility and autonomy make constraints central.}
In read-mostly or reversible digital settings, recovery can often be handled through retries and checkpoints.
In grounding-dominant web interaction, safety-dominant medical assistance, and control-dominant physical settings, actions can have persistent side effects.
Verification and Governance therefore becomes part of the primary bottleneck rather than a peripheral add-on.
Higher autonomy magnifies this pattern because the harness must absorb responsibilities that a human operator would otherwise carry.

\paratitle{Long-horizon performance depends on externalized state across all domains.}
Whether the task is repo-scale coding (L3), literature synthesis (L3-L4), or embodied exploration (L4), one prompt window is rarely the right unit of memory.
Durable artifacts, summaries, checkpoints, plans, and logs keep trajectories coherent over time.
The configuration consequence is that the Context Manager and State and Artifact Store must be designed jointly: summaries decide what is visible now, whereas artifacts and checkpoints decide what remains recoverable later.

\paratitle{Task pressure should be reported together with evaluation results.}
The mapping in Tab.~\ref{tab:task-harness-mapping} also constrains benchmark interpretation.
Benchmarks are most informative when they stress the harness components that match the primary bottleneck of a target deployment setting.
Benchmark reports should therefore describe not only the model, but also the task pressures and harness configuration under which results are obtained.

\section{Evaluation and Empirical Analysis}
\label{sec:sec7:eval}

Evaluation makes the model--harness interaction directly observable.
Benchmark scores should therefore be interpreted as outcomes of a \emph{model--harness pairing}: the same model may behave differently under different context policies, tool interfaces, control loops, verification procedures, and retry budgets.
This section uses representative benchmarks to test this view across three interaction regimes: software-engineering tasks with strong test oracles, terminal tasks with command-line execution and environment manipulation, and web tasks with browser grounding and stateful interaction.
Across these regimes, we try to answer three questions: \emph{how much performance is explained by stronger backbone models, how much variation remains after conditioning on the model, and how runtime cost, latency, timeout behavior, and trace availability change the interpretation of task success}.

\subsection{Benchmark Landscape and Evaluation Work}
Existing evaluation work can be organized as a pipeline that turns an agent run into interpretable evidence: benchmarks specify the task, execution infrastructures standardize the run, judgment methods score and diagnose the outcome, and continuous evaluation practices feed these signals back into system improvement.

\noindent\textbf{Benchmarks as task specifications.}
Tab.~\ref{tab:benchmarks} summarizes representative benchmarks for LLM-based agents.
Beyond application domains, these benchmarks stress different harness capabilities: SWE-bench~\cite{jimenez2024swe} tests repository navigation, code editing, and hidden-test verification; WebArena~\cite{zhou2024webarena}, VisualWebArena~\cite{koh2024visualwebarena}, and OSWorld~\cite{xie2025osworld} test web/GUI grounding, state tracking, multimodal perception, and safe interface control; Terminal-Bench~\cite{merrill2026terminalbench} tests command-line execution and environment manipulation; and LoCoMo~\cite{maharana2024locomo} and OS-Harm~\cite{kuntz2025osharm} test memory persistence and harmful-action control.
Together, they mark a shift toward ecologically realistic evaluation, where browsers, terminals, and operating systems reveal long-horizon failures that closed-form datasets often miss.

\noindent\textbf{Execution and trace infrastructure.}
For coding and terminal agents, systems such as SWE-agent~\cite{yang2024sweagent}, OpenHands~\cite{wang2024openhands}, Repo2Run~\cite{hu2026repo2run}, R2E-Gym~\cite{jain2025r2e}, and HAL~\cite{kapoor2025holistic} use controlled environments, sandboxed execution, standardized rollouts, and trace collection to make evaluation reproducible and diagnosable.
This infra helps distinguish harness behavior from artifacts of dependency drift, invalid graders, changed tool interfaces, or inconsistent resets. Representative harness designs are summarized in Tab.~\ref{tab:harness-designs}.
 
\noindent\textbf{Judgment and attribution methods.}
Executable tests and state-based checkers provide strong oracles for coding and terminal tasks, whereas open-ended outputs often require LLM-as-judge or human-audit protocols.
G-Eval~\cite{liu2023g}, MT-Bench~\cite{zheng2023judging}, and surveys of LLM-as-a-judge~\cite{gu2024survey} show the promise and risks of model-based evaluators, including bias, inconsistency, and evaluator drift.
For harness engineering, judgment matters most when it attributes failures to model reasoning, context construction, tool exposure, execution control, safety constraints, or the evaluator itself.

\noindent\textbf{Continuous evaluation practices.}
Frameworks such as LangChain's agent evaluation tooling~\cite{trivedy2026deepevals}, DeepEval~\cite{deepeval2025}, RAGAS~\cite{es2024ragas}, and lm-evaluation-harness~\cite{gao2021framework} support recurring tests, trace inspection, judge-based metrics, and monitoring-style evaluation.
Their role is not merely to report a leaderboard score, but to make evaluation reusable during prompt changes, tool updates, context-policy revisions, and deployment monitoring.

\begin{table}[t]
\centering
\caption{\small{Representative benchmarks for LLM-based agents.}}
\label{tab:benchmarks}
\scriptsize
\setlength{\tabcolsep}{2pt}
\resizebox{\columnwidth}{!}{
\begin{tabular}{@{}p{2.6cm}p{1.9cm}p{2.2cm}p{2.4cm}@{}}
\toprule
\textbf{Benchmark} & \textbf{Focus} & \textbf{Environment} & \textbf{Primary metric} \\
\midrule
AgentBench~\cite{liu2023agentbench} & General & Interactive envs & Task completion \\
SWE-bench~\cite{jimenez2024swe} & Coding & Real GitHub issues & Resolution rate \\
WebArena~\cite{zhou2024webarena} & Web & Realistic websites & Task success \\
VisualWebArena~\cite{koh2024visualwebarena} & Multimodal web & Visual web tasks & Task success \\
OSWorld~\cite{xie2025osworld} & Desktop & Real OS & Multi-app success \\
Terminal-Bench~\cite{merrill2026terminalbench} & Terminal/Coding & Command-line & Task success \\
MCPWorld~\cite{yan2025mcpworld} & API+GUI & Hybrid tool envs & Task success/tool use \\
OS-Harm~\cite{kuntz2025osharm} & Safety & Desktop computer & Harmful action rate \\
LoCoMo~\cite{maharana2024locomo} & Long-term mem. & Multi-session chat & QA/consistency \\
MMAU~\cite{yin2024mmau} & General & Cross-domain & Capability scores \\
MLE-Bench~\cite{chan2024mlebench} & ML engineering & Kaggle-like tasks & Performance tier \\
% MCPAgentBench~\cite{liu2025mcpagentbench} \\
% MCP-Atlas~\cite{bandi2026mcpatlas} \\
% GAIA~\cite{mialon2024gaia} \\
MCPAgentBench~\cite{liu2025mcpagentbench} & MCP tool use & MCP sandbox & Task Compl./eff. \\
MCP-Atlas~\cite{bandi2026mcpatlas} & MCP tool use & Real MCP servers & Pass rate  \\
GAIA~\cite{mialon2024gaia} & General assistant & Web/files/tools & Answer accuracy \\
Claw-SWE-Bench~\cite{claw-swe-bench} & Agent harnesses & Real GitHub issues & Resolution rate/cost \\
TheAgentCompany~\cite{xu2026theagentcompany} & Enterprise-style & Simulated company & Task success \\
\bottomrule
\end{tabular}
}
\end{table}

\subsection{Evaluation Dimensions Beyond Task Success}
\label{sec:eval:dimensions}

Most public agent leaderboards still rank systems by a single outcome-centric score.
SWE-bench~\cite{jimenez2024swe} reports the percentage of resolved issues~\cite{swebenchleaderboard}, while Terminal-Bench~\cite{merrill2026terminalbench} primarily report task-level completion scores~\cite{terminalbenchleaderboard}.
However, recent evaluation studies~\cite{yang2024sweagent,pan2026nlah,lee2026metaharness} increasingly argue that accuracy alone is insufficient for agent assessment.
For example, Sayas~\etal~\cite{kapoor2024ai} calls for cost-controlled and reproducible evaluation. CLEAR~\cite{mehta2025beyond} explicitly evaluates cost, latency, efficacy and assurance. ReliabilityBench~\cite{gupta2026reliabilitybench} studies consistency, robustness, and fault tolerance under production-like stress. Procedure-aware evaluation~\cite{cao2026beyond} shows that apparent task completion can hide unsafe or invalid trajectories.
Because an agent run couples model reasoning, harness design, environment setup, tool interfaces, and evaluator logic, a failure may originate from any part of this chain.
Task success remains the primary outcome metric, but \textbf{meaningful harness comparison requires a richer reading of results along several additional dimensions:}
\begin{itemize}[leftmargin=*]
  \item \textbf{Task success:} whether the final objective is completed.
  \item \textbf{Reliability:} whether performance remains stable across stochastic runs, retries, and environment variations.
  \item \textbf{Efficiency:} token usage, API cost, and compute cost.
  \item \textbf{Latency:} wall-clock time or number of interactions.
  \item \textbf{Safety:} whether actions remain within allowed boundaries and avoid harmful side effects.
  \item \textbf{Process quality:} whether the trajectory is inspectable, recoverable, and evidence-backed.
\end{itemize}

These dimensions explain why similar final scores can hide substantial harness differences.
One harness may trade long trajectories, repeated retries, and heavy context accumulation for higher success, while another may deliver slightly lower success at much lower cost and latency.
From a deployment perspective, the key is not raw success alone, but useful task completion under resource constraints.

In the following empirical analyses, we focus on the dimensions that are most consistently available across public reports and leaderboard logs: task success, runtime, timeout behavior, and token usage when available.
Monetary cost is discussed only cautiously because public cost fields are sparse and often depend on harness-specific accounting, cache handling, and model-price assumptions.

\subsection{Harness Effects on SWE-bench Verified}
\label{sec:eval:swebench}

SWE-bench Verified~\cite{jimenez2024swe} comprises 500 human-validated GitHub issues drawn from twelve Python repositories, each paired with a hidden test suite that provides a deterministic pass/fail oracle.
Evaluations run inside sandboxed Docker containers with pinned dependencies, so observed differences reflect system capabilities rather than environmental artifacts.
By filtering out ambiguous specifications and broken tests from the original 2{,}294-instance SWE-bench, the Verified split offers a cleaner signal for cross-system comparison.
Solving an instance requires localizing the defect, editing source files, and passing hidden regression tests; because different harnesses partition these stages in distinct ways, the benchmark is well suited for studying how scaffold design affects measured performance.

\begin{table}[H]
\centering
\caption{\small Model--harness results on SWE-bench Verified. Resolved rates are percentages, resolved counts are out of 500 instances; cost is USD per instance when reported; vendor rows are proprietary references.}
\label{tab:swebench_harness_model}
\footnotesize
\setlength{\tabcolsep}{3pt}
\renewcommand{\arraystretch}{0.82}
\resizebox{0.98\columnwidth}{!}{
\begin{tabular}{@{}llcc@{}}
\toprule
\textbf{Primary model} & \textbf{Harness / scaffold} & \makecell{\textbf{Res. (\%)}\\\textbf{/ solved}} & \makecell{\textbf{Cost}\\\textbf{(\$)}} \\
\midrule
\multirow{3}{*}{GPT-4o} & SWE-agent & 23.2 / 116 & - \\
& AutoCodeRover-v2 & 38.4 / 192 & - \\
& Agentless & 38.8 / 194 & - \\
\midrule
\multirow{6}{*}{Claude 3.5 Sonnet} & SWE-agent (20240620) & 33.6 / 168 & - \\
& SWE-agent + tools & 49.0 / 245 & - \\
& Agentless & 50.8 / 254 & 1.19 \\
& AutoCodeRover & 51.8 / 259 & 4.50 \\
& OpenHands + CodeAct 2.1 & 53.0 / 265 & 0.78 \\
& PatchPilot & 53.6 / 268 & 0.99 \\
\midrule
\multirow{3}{*}{Claude 3.7 Sonnet} & mini-SWE-agent & 52.8 / 264 & - \\
& SWE-agent + tools & 63.2 / 316 & - \\
& Vendor scaffold & \emph{63.7} / - & - \\
\midrule
\multirow{4}{*}{Claude Sonnet 4} & mini-SWE-agent & 64.9 / 325 & - \\
& OpenHands + CodeAct 2.1 & 70.4 / 352 & - \\
& SWE-agent + tools & 72.4 / 362 & - \\
& Vendor scaffold & \emph{72.7} / - & - \\
\midrule
\multirow{4}{*}{Claude Opus 4 / 4.5} & SWE-agent + tools & 73.2 / 366 & - \\
& mini-SWE-agent & 76.8 / 384 & - \\
& OpenHands + CodeAct 2.1 & 77.6 / 388 & - \\
& Vendor scaffold & \emph{80.9} / - & - \\
\midrule
o3 / o4-mini & PatchPilot v1.1 & 64.6 / 323 & - \\
\midrule
\multirow{3}{*}{OpenAI GPT-5 family} & OpenHands + CodeAct 2.1 & 71.8 / 359 & - \\
& mini-SWE-agent & 72.8 / 364 & - \\
& Vendor scaffold & \emph{80.0} / - & - \\
GPT-5.3 Codex & mini-SWE-agent & 78.0 / 390 & - \\
GPT-5.4 & mini-SWE-agent & 78.2 / 391 & - \\
GPT-5.5 & mini-SWE-agent & 82.6 / 413 & - \\
\midrule
\multirow{2}{*}{Gemini 3 Pro} & mini-SWE-agent & 74.2 / 371 & - \\
& Vendor scaffold & \emph{76.2} / - & - \\
Gemini 3.1 Pro Preview & mini-SWE-agent & 78.8 / 394 & - \\
\midrule
\multirow{2}{*}{DeepSeek V3 / V3.2} & Agentless & 42.0 / 210 & - \\
& mini-SWE-agent & 70.0 / 350 & - \\
Claude Opus 4.6 Thinking & mini-SWE-agent & 78.2 / 391 & - \\
Claude Opus 4.7 & mini-SWE-agent & 82.0 / 410 & - \\
\bottomrule
\end{tabular}
}
\begin{minipage}{0.99\linewidth}
\vspace{1pt}
\scriptsize{
\textit{Notes.} Claude 3.5 Sonnet entries refer to the 2024-10-22 snapshot where specified; Claude Sonnet 4 and Opus 4 source-card entries use the 2025-05-14 generation. Some rows differ in inference policy, including extended-thinking or high-reasoning settings. Vendor rows use closed-source scaffolds.}
\end{minipage}
\end{table}

Tab.~\ref{tab:swebench_harness_model} compares SWE-bench Verified results across several model--harness pairings, including open-source scaffolds, source-reported agent harnesses, lightweight mini-SWE-agent runs, and closed-source vendor reports.
The table should be read as a compact synthesis of public evidence rather than a fully controlled factorial experiment.
Model snapshots, reasoning settings, retry budgets, and proprietary scaffold details are not always aligned across sources, so the strongest comparisons are those within the same row family or under the same reported evaluation setting.
Vendor-reported scores are included as upper-envelope references, but they should not be interpreted as controlled ablations against open-source harnesses.
For example, Opus 4.5 with mini-SWE-agent is reported with extended thinking, whereas the same source reports 74.4\% under medium reasoning and 67.6\% for Opus 4; the OpenAI mini-SWE-agent row follows a GPT-5-2 extended-thinking style leaderboard setting, whereas GPT-5 (2025-08-07) under medium reasoning is reported at 65.0\%, and the vendor 80.0\% reference is a GPT-5.2 result reported in a Claude system card~\cite{swebenchleaderboard,anthropic2026sonnet46sc}.
Similarly, DeepSeek rows distinguish V3 from V3.2 high-reasoning settings, and Gemini rows distinguish Gemini~3~Pro from later Gemini~3.1~Pro reports, including an 80.6\% Gemini~3.1~Pro result under its own reported configuration~\cite{google2026gemini31mc}.
These boundaries do not invalidate the table, but they mean the strongest claims should use within-row-family comparisons and treat vendor or reasoning-policy changes as upper-envelope evidence rather than controlled ablations.

The compared harnesses span a broad spectrum of scaffold complexity.
Agentless~\cite{xia2025demystifying} removes the interactive agent loop and uses a fixed localize--repair--validate pipeline.
SWE-agent + tools~\cite{anthropic2025swebench,yang2024sweagent} exposes shell and editing tools through a bash-oriented repair loop, while mini-SWE-agent~\cite{minisweagent2025} reduces this design to a minimal scaffold that leaves most orchestration to the model.
OpenHands + CodeAct~2.1~\cite{wang2024openhands} provides a richer software-engineering runtime with file editing, web browsing, and IPython execution.
AutoCodeRover~\cite{zhang2024autocoderover} and PatchPilot~\cite{li2025patchpilot} represent more structured repair workflows, using repository search, localization, reproduction, validation, and refinement to constrain the repair process.
Vendor scaffold lists the best vendor-reported scores on proprietary scaffolds~\cite{anthropic2025claude37,anthropic2025claude4,anthropic2026sonnet46sc,openai2025gpt5,google2025gemini3blog} as an upper-envelope reference.
Together, these systems provide a useful, though not perfectly controlled, view of how scaffold design interacts with backbone model capability on repository-level coding tasks.

Specifically, \textbf{model capability and harness design both contribute to measured performance.}
Within a single harness, backbone upgrades drive large gains: SWE-agent + tools improves from 49.0\% with Claude~3.5~Sonnet to 73.2\% with Opus~4, a 24\% increase~\cite{anthropic2025swebench,swebenchexperiments}; mini-SWE-agent shows a comparable trajectory from 52.8\% (Claude~3.7~Sonnet) to 76.8\% (Opus~4.5)~\cite{swebenchleaderboard}.
Within a single model, harness choice also produces a consistent effect.
For GPT-4o, source-reported harnesses range from 23.2\% with SWE-agent to 38.8\% with Agentless.
For Claude~3.5~Sonnet, they range from 33.6\% with SWE-agent to 53.6\% with PatchPilot, with SWE-agent + tools, Agentless, AutoCodeRover, and OpenHands + CodeAct~2.1 occupying the middle of the range.
For Claude Opus~4/4.5, the spread is narrower but still visible: 73.2\% with SWE-agent + tools, 76.8\% with mini-SWE-agent, and 77.6\% with OpenHands + CodeAct~2.1.
These ranges show that the same backbone can gain or lose tens of resolved instances depending on the scaffold.

\textbf{Scaffold complexity does not predict effectiveness.}
Under Opus~4.5, mini-SWE-agent (roughly 100 lines of Python) reaches 76.8\%, only slightly below the far richer OpenHands + CodeAct~2.1 sandbox at 77.6\%~\cite{swebenchleaderboard,swebenchexperiments}.
These results suggest that scaffold effectiveness depends more on interface design than on feature count: a minimal scaffold with well-chosen primitives can extract nearly the same performance as a full-featured agent framework.

Vendor-reported scores, which reflect proprietary scaffold optimization, consistently exceed the best open-source results.
OpenHands with Opus~4.5 at 77.6\%~\cite{swebenchexperiments} trails the corresponding vendor score of 80.9\%~\cite{anthropic2026sonnet46sc} by about 3\%; for Gemini~3~Pro, the gap narrows to 2\% (74.2\% \vs 76.2\%)~\cite{google2025gemini3blog,minisweagent2025}.
For GPT-5 variants the margin is larger (72.8\% \vs 80.0\%), though differences in model version and inference configuration complicate this comparison~\cite{anthropic2026sonnet46sc,swebenchleaderboard}.
Across same-generation Claude and Gemini models, this 2-4\% advantage is attributable to scaffold-level decisions such as prompt design, candidate selection, and compute scaling, not to differences in model capability.

\subsection{Harness Effects on Terminal-Bench 2.0}
\label{sec:eval:terminalbench}
Terminal-Bench provides a complementary perspective to SWE-bench because the agent must operate through an interactive command-line environment rather than only submit a final patch~\cite{merrill2026terminalbench}.
Each task specifies a natural-language instruction, a sandboxed terminal workspace, an executable test script, and a reference solution, so success is defined by whether the agent transforms the environment into a passing state.
Tasks commonly require file inspection, tool installation or invocation, command execution, log interpretation, artifact editing, and explicit termination decisions.
The benchmark is therefore well suited for studying execution harnesses, since terminal interaction jointly exercises observation design, context management, control-loop policy, action exposure, state persistence, and verification.

Our analysis uses the official Terminal-Bench 2.0 leaderboard and public submission logs as data sources~\cite{terminalbenchleaderboard,terminalbenchsubmissions}.
Official submissions evaluate \texttt{terminal-bench@2.0} with five trials per task (\texttt{-k 5}), use each task's benchmark environment and default constraints, and must not override timeouts or CPU, memory, and storage limits~\cite{terminalbenchsubmissions}.
Leaderboard-integrity rules further penalize reward-hacking trajectories, such as retrieving task solutions from the internet, which reduces the risk that reported scores reflect benchmark leakage rather than terminal task completion~\cite{terminalbenchintegrity}.
For the performance analysis, we use entries for which the backbone model and harness are identifiable, excluding entries whose model field is a mixture or ``Multiple'' so that each plotted point has a clear model identity.
The resulting evidence is not a randomized ablation, since public submissions may differ in prompts, versions, budgets, and implementation details.
It is nevertheless informative as an observational comparison: the same model appears under several harnesses, and the same harness often appears with several models.
For resource analysis, we use the official HuggingFace public-submission repository as the primary source and align submissions to the currently visible leaderboard by strict metadata matching.
The public repository contains 75 submissions with metadata and logs, covering 32{,}604 trial records.
Among these, 48 submissions strictly match a currently visible leaderboard entry.
Reward, agent-runtime, and full-runtime fields have high coverage (97.2\%, 98.1\%, and 100.0\%, respectively), whereas input/output token fields cover 45.0\% of trial records and dollar-cost fields cover only 15.2\%.
Accordingly, runtime and timeout statistics are the main resource-efficiency evidence, while token statistics are used as supplementary evidence and monetary cost is not used for cross-harness claims.
These counts define the evidence boundary rather than a separate result table: 27 public submissions are excluded from the complete resource table because 24 are not visible or name-mismatched on the current leaderboard and 3 are ambiguous matches.
A looser visible-entry check finds 55 matches among 142 visible leaderboard entries (38.7\%) across 51 unique submissions, but it is used only as a coverage sanity check.
The row-level comparisons below use the strict 48-row subset.
Tab.~\ref{tab:tb2_efficiency} extracts representative same-model rows from this subset for the main-text comparison.

\begin{table}[t]
\centering
\caption{\small{Terminal-Bench 2.0 representative resource statistics for leaderboard-visible submissions. Input/Output report median token counts in thousands per trial, Agent reports median runtime in minutes, and TO reports timeout rate.}}
\label{tab:tb2_efficiency}
\scriptsize
\setlength{\tabcolsep}{1.8pt}
\renewcommand{\arraystretch}{0.94}
\resizebox{\columnwidth}{!}{%
\begin{tabular}{@{}llrrrrrr@{}}
\toprule
\textbf{Model} & \textbf{Harness}
& \makecell{\textbf{Score}\\\textbf{(\%)}}
& \textbf{Trials}
& \makecell{\textbf{Input}\\\textbf{(K)}}
& \makecell{\textbf{Output}\\\textbf{(K)}}
& \makecell{\textbf{Agent}\\\textbf{(min)}}
& \makecell{\textbf{TO}\\\textbf{(\%)}} \\
\midrule
\multirow{3}{*}{GPT-5.3 Codex}
 & SageAgent & 78.4 & 445 & - & - & 5.7 & 12.1 \\
 & Mux & 74.6 & 445 & 238.7 & 5.7 & 5.5 & 8.1 \\
 & Terminus 2 & 64.7 & 445 & 58.4 & 20.5 & 8.9 & 20.7 \\
\midrule
\multirow{4}{*}{Claude Opus 4.6}
 & Meta-Harness & 76.4 & 445 & 755.0 & 15.8 & 6.3 & 7.9 \\
 & Terminus-KIRA & 74.7 & 445 & 618.6 & 16.9 & 9.6 & 18.2 \\
 & Mux & 66.5 & 445 & 213.0 & 10.1 & 5.9 & 10.3 \\
 & Terminus 2 & 62.9 & 445 & 79.4 & 8.4 & 5.3 & 18.2 \\
\midrule
\multirow{2}{*}{Gemini 3.1 Pro}
 & TongAgents & 80.2 & 445 & - & - & 11.1 & 21.1 \\
 & Terminus-KIRA & 74.8 & 445 & 257.5 & 22.1 & 6.5 & 9.7 \\
\bottomrule
\end{tabular}%
}
\end{table}

\begin{figure*}[t]
\centering
\includegraphics[width=\textwidth]{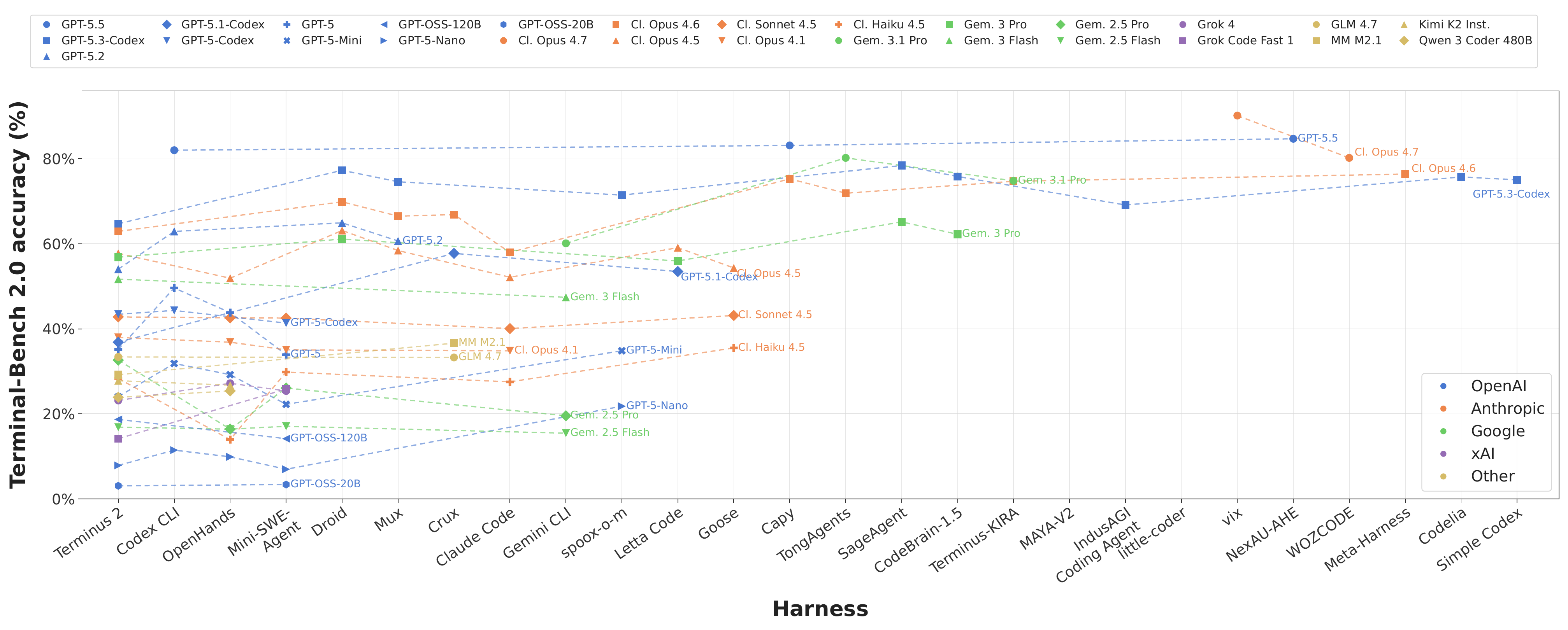}
\vspace{-18pt}
\caption{\small{Terminal-Bench 2.0 accuracy across model--harness pairings. Each point is a single-backbone leaderboard entry, and dashed lines connect results that use the same model under different harnesses.}}
\label{fig:tb2_arc_harness_effects}
\end{figure*}

\begin{figure*}[t]
\centering
\includegraphics[width=\textwidth]{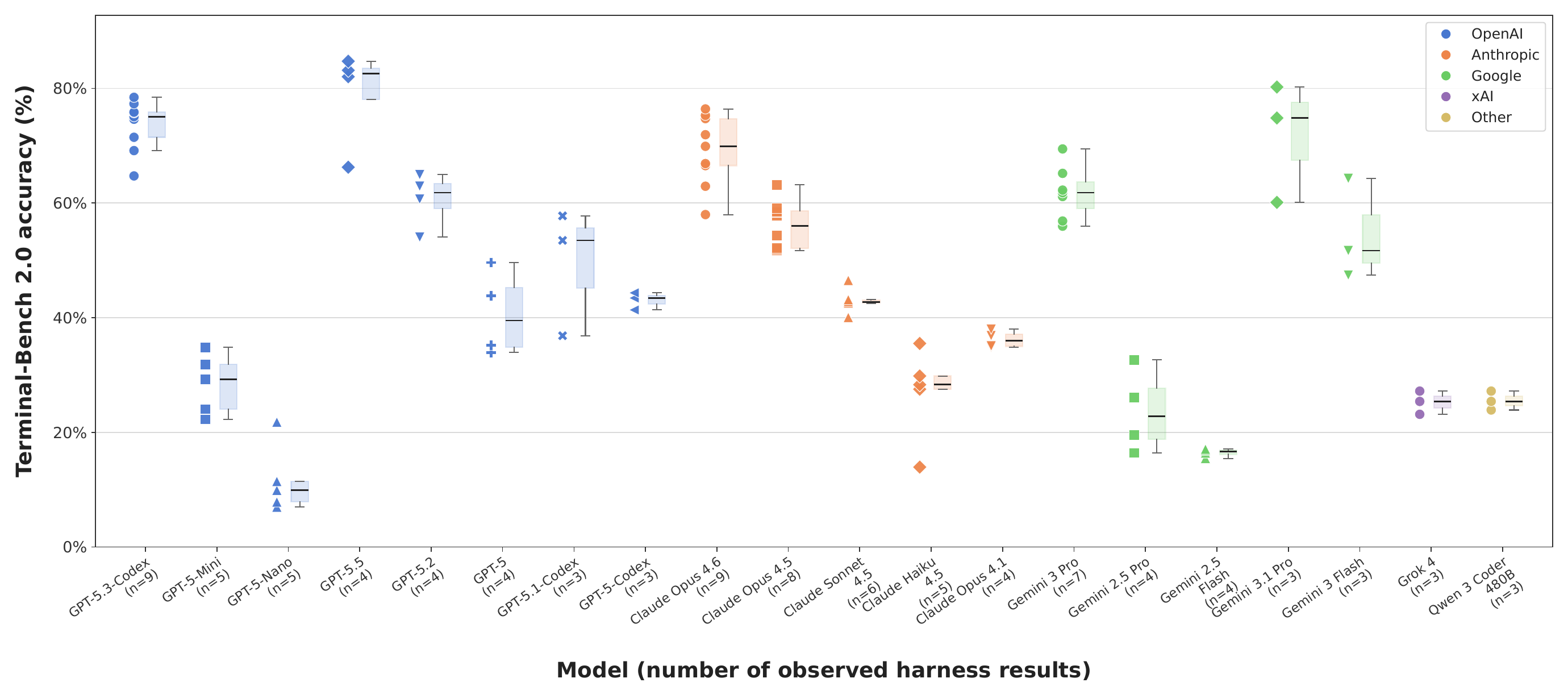}
\vspace{-18pt}
\caption{\small{Within-model variation on Terminal-Bench 2.0. For each model with at least three observed harness results, the box and points summarize accuracy across harnesses.}}
\label{fig:tb2_model_harness_distribution}
\end{figure*}

Fig.~\ref{fig:tb2_arc_harness_effects} first indicates that \textbf{backbone capability remains a major determinant of success.}
Within a fixed harness, stronger model generations often improve resolved rates by more than 10\%.
For example, under Terminus 2, GPT-5 improves from 35.2\% to 54.0\% with GPT-5.2 and to 64.7\% with GPT-5.3-Codex.
Under Codex CLI, the sequence from GPT-5 to GPT-5.2 and GPT-5.5 rises from 49.6\% to 62.9\% and 82.2\%.
The same pattern appears for Anthropic and Google models: newer Opus, Gemini, and Codex-specialized models generally occupy higher regions of the plot than earlier or smaller models.
Thus, the benchmark does not support a harness-only interpretation; terminal agents still need strong planning, coding, debugging, and tool-use priors from the foundation model.

At the same time, conditioning on the model reveals large harness-induced spreads.
GPT-5.3-Codex ranges from 64.7\% with Terminus 2 to 78.4\% with SageAgent, a 13.7\% difference.
Claude Opus 4.6 ranges from 58.0\% with Claude Code to 76.4\% with Meta-Harness, an 18.4\% difference.
Gemini 3.1 Pro ranges from 59.4\% with Gemini CLI to 80.2\% with TongAgents, a 20.8\% difference.
Even GPT-5, before the later Codex-specialized variants, ranges from 33.9\% with Mini-SWE-Agent to 49.6\% with Codex CLI.
These gaps are substantially larger than the standard errors reported for most relevant leaderboard entries, and they correspond to different harness choices around terminal affordances, context packaging, command execution, stopping criteria, and recovery.

Fig.~\ref{fig:tb2_model_harness_distribution} aggregates this fixed-model view.
Among the 20 models that have at least three observed harness results, the median within-model range is 13.6\% and 14 of the 20 models vary by at least 10\% across harnesses.
The largest spreads exceed 20\%, as seen for Claude Haiku 4.5, GPT-5.1-Codex, and Gemini 3.1 Pro.
This distribution shows that leaderboard accuracy cannot be attributed to the model alone.
A model's measured terminal competence also depends on whether the harness exposes an effective command interface, preserves relevant execution state, routes observations back into context at an appropriate granularity, and uses verification signals for termination, retry, or repair.

Tab.~\ref{tab:tb2_efficiency} shows that accuracy differences are accompanied by distinct operational profiles.
For GPT-5.3-Codex, SageAgent reaches the highest single-backbone score in Fig.~\ref{fig:tb2_arc_harness_effects} with a median agent time of 5.7 minutes and a 12.1\% timeout rate, whereas Terminus~2 takes 8.9 minutes and times out on 20.7\% of trials.
Mux uses a heavier median input context than Terminus~2 (238.7K versus 58.4K tokens) but reports shorter median agent time (5.5 versus 8.9 minutes) and a lower timeout rate (8.1\% versus 20.7\%).
For Claude Opus~4.6, Meta-Harness uses a much larger median input context than Terminus~2 (755.0K versus 79.4K tokens) while reducing timeout rate from 18.2\% to 7.9\%.
These examples show that the measured system behavior includes not only whether a task is solved, but also how much context is consumed, how long execution takes, and how often the harness fails to terminate successfully.

Taken together, Terminal-Bench results separate two effects without treating either as sufficient on its own.
Model upgrades lift whole harness families, but harness choices can still shift the same model's score by more than 10\% through terminal-state presentation, context management, command execution, and verifier feedback.
Resource statistics add a deployment-facing caveat: higher accuracy may require longer trajectories, heavier context, or more robust timeout handling, and these costs are part of the practical capability being measured.
On Terminal-Bench, reliable terminal task completion therefore depends on the fit between the model's interactive skills and the harness's runtime design under fixed benchmark constraints.

\subsection{Harness Effects on WebArena}
\label{sec:eval:webarena}

WebArena~\cite{zhou2024webarena} is one of the most widely used reproducible benchmarks for web agents.
It evaluates agents in self-hosted websites that cover realistic domains such as shopping, discussion forums, GitLab-style software collaboration, content management, maps, and knowledge resources.
Unlike open-ended browsing benchmarks that often require human or LLM-based judgement, WebArena uses programmatic success checks over website state and task-specific answers.
This makes it useful for studying what a web-agent harness contributes beyond a model-only baseline: browser agents must convert textual goals and visual or DOM observations into navigation, search, form filling, state tracking, and recovery actions, while the evaluator supplies a relatively concrete task-success signal.

\begin{table}[t]
\centering
\caption{\small WebArena task-success evidence by backbone. Scores are percentages; Span reports the high--low difference in percentage points for each backbone.}
\label{tab:webarena_excel_full}
\footnotesize
\setlength{\tabcolsep}{3pt}
\renewcommand{\arraystretch}{0.86}
\resizebox{\columnwidth}{!}{
\begin{tabular}{@{}lcccc@{}}
\toprule
\textbf{Backbone}
& \textbf{Model only}
& \textbf{Harness low}
& \textbf{Harness high}
& \textbf{Span}\\
\midrule
GPT-3.5 & 8.9 & 22.0 & 29.1 & 20.2 \\
GPT-4 & 14.9 & 20.2 & 33.0 & 18.1 \\
GPT-4o & 13.1 & 19.2 & 54.6 & 41.5 \\
GPT-4 Turbo & 16.5 & 33.3 & 45.7 & 29.2 \\
GPT-4o-mini & - & 13.6 & 13.6 & - \\
GPT-5 & - & 71.2 & 71.2 & - \\
DeepSeek R1-Llama 8B & 8.5 & 43.6 & 43.6 & 35.1 \\
DeepSeek V3.2 & - & 74.3 & 74.3 & - \\
Gemini 3 Pro & - & 51.2 & 71.6 & 20.4 \\
Gemini 3.1 Flash-L & - & 42.3 & 42.3 & - \\
Claude Sonnet 3.5 & - & 36.2 & 52.1 & 15.9 \\
Qwen3.5 family & - & 3.1 & 41.5 & 38.4 \\
Llama 3-70B & 7.6 & 10.1 & 10.1 & 2.5 \\
Llama 3.1-70B & - & 18.4 & 18.4 & - \\
Llama 3.2-1B & 2.4 & 24.1 & 24.1 & 21.7 \\
Llama 3.1-8B & 5.6 & 48.5 & 48.5 & 42.9 \\
\bottomrule
\end{tabular}
}
\end{table}

Tab.~\ref{tab:webarena_excel_full} summarizes sparse WebArena results as backbone-level evidence slices, including both model-only references and harnessed agent systems.
Rows containing the same backbone under multiple settings provide the strongest evidence for harness effects.
Because most public WebArena reports describe complete agent systems rather than controlled factorial ablations, differences in prompts, browser actions, observation formats, retry policies, search budgets, training data, and implementation details should be treated as part of the reported system configuration~\cite{zhou2024webarena,webarenaleaderboard2026,lesellier2025browsergym}.

The clearest harness effects come from backbones that have both model-only and harnessed results.
GPT-4o ranges from 13.1\% in the model-only baseline to 54.6\% with WebOperator, a 41.5\% span; even among named harnesses alone, it ranges from 19.2\% with LM-TS to 54.6\% with WebOperator~\cite{koh2024treesearch,dihan2025weboperator}.
GPT-4 improves from 14.9\% to 33.0\% with SteP, GPT-4-Turbo from 16.5\% to 45.7\% with AgentOccam, and GPT-3.5 from 8.9\% to 29.1\% under the stronger WebPilot entry~\cite{sodhi2023step,yang2024agentoccam,zhang2024webpilot}.
The same phenomenon appears for open-weight or distilled models: DeepSeek-R1-Distill-Llama-8B moves from 8.5\% to 43.6\% with AgentSymbiotic, Llama-3.2-1B from 2.4\% to 24.1\%, and Llama-3.1-8B from 5.6\% to 48.5\%~\cite{zhang2025agentsymbiotic}.
These gaps are too large to be explained by task noise alone; they reflect how observation design, search, workflow memory, action grounding, and stopping policies transform a language model into an effective web actor.
Futhermore, WebTactix with DeepSeek V3.2 reaches 74.3\%, corresponding to 594 solved tasks out of 812 in its public report~\cite{webtactix2026}.
OpAgent reaches 71.6\% with Gemini 3 Pro, and ColorBrowserAgent reaches 71.2\% with GPT-5~\cite{guo2026opagent,wang2026colorbrowseragent}, showing that recent web-agent systems can exceed the 70\% level on WebArena, but they combine strong backbones with specialized runtime structure, grounding, search, and adaptive memory; they should therefore be treated as model--harness system results.

Fixed-harness comparisons show the complementary role of model capability.
BrowserGym~\cite{lesellier2025browsergym} provides the broadest same-harness slice: scores range from 51.2\% for Gemini~3~Pro to 42.3\% for Gemini~3.1~Flash-L, 41.5\% for Qwen3.5-27B, 36.2\% for Claude~3.5~Sonnet, 31.4\% for GPT-4o, 23.5\% for GPT-4, 18.4\% for Llama-3.1-70B, and 13.6\% for GPT-4o-mini~\cite{lesellier2025browsergym}.
Within Qwen3.5, performance falls monotonically from 27B to 9B, 4B, and 2B, indicating that stronger backbones generally improve planning, instruction following, and state tracking under a common browser interface.
Yet model size does not fully determine outcomes: GPT-4o under BrowserGym trails Gemini~3.1~Flash-L and Qwen3.5-27B, while AgentSymbiotic with Llama-3.1-8B exceeds several larger-model BrowserGym results. The same backbone can be under-expressed by one harness and amplified by another.

Overall, WebArena reinforces the model--harness view from a web-interaction setting.
In coding benchmarks, the harness shapes how tests, edits, and repository context are exposed; in WebArena, it shapes how a model sees the page, chooses browser actions, recovers from navigation errors, and verifies that a website state has changed as intended.
Programmatic scoring reduces evaluator drift compared with LLM-as-judge protocols, but it does not remove all measurement risk: brittle checkers, ambiguous instructions, and environment leakage can still distort results.
For high-confidence claims, audited variants such as WebArena Verified~\cite{webarenaverified} are preferable when available because they retain the reproducible WebArena environment while repairing evaluator and instruction artifacts.
Consequently, browser-agent success is best reported as a conditional property of a complete model--harness system, together with its observation mode, action interface, search or retry budget, memory policy, and source artifacts.

\subsection{Benchmark Insights}
\label{sec:eval:insights}

Several lessons emerge from comparing harness behavior across benchmarks.

\noindent\textbf{Harness design should match the benchmark oracle.}
When benchmark provides strong automatic feedback, as in coding tasks with tests, effective harnesses exploit verifier loops, patch refinement, and rollback.
Terminal-Bench reinforces the same principle in command-line environments: useful harnesses turn command output, files, and completion checks into actionable feedback for termination, retry, and repair.
When the oracle is weak or delayed, as in research or workplace tasks, the harness must rely more on provenance tracking, intermediate review, and conservative stopping.

\noindent\textbf{Autonomy and complexity are not monotonic goods.}
Fully open-ended loops explore broadly but can accumulate context, drift, and cost.
When objectives are narrow and success is mechanically checkable, structured pipelines such as localization--repair--validation can outperform more agentic loops, and compact scaffolds can match richer runtimes.
The key design question is not how much autonomy the harness exposes, but which degrees of freedom help the model exploit the benchmark's feedback structure.

\noindent\textbf{Model--harness compatibility matters.}
A strong model may perform poorly under a harness that exposes the wrong action space or overloads the context window.
Conversely, a lightweight scaffold can be effective when it matches the model's preferred interaction pattern and the benchmark's feedback structure.
On Terminal-Bench 2.0, the same model can vary by double-digit accuracy across harnesses; on WebArena, the gap between model-only baselines and browser-agent scaffolds can exceed 40\% for GPT-4o.
These differences make compatibility an empirical property of the model--harness pair rather than an implementation detail.

\noindent\textbf{Scores are conditional on runtime configuration.}
Tool privileges, context policy, retry budget, sandbox restrictions, and completion criteria all shape measured performance.
The Terminal-Bench public logs further show that runtime and timeout profiles vary substantially even among leaderboard-visible submissions.
Thus, a benchmark score is interpretable only together with the runtime configuration that produced it.
Reports should include at least model version, harness identity, tool privileges, retry and timeout policy, execution environment, token or API usage when available, and trace or verifier metadata.

\noindent\textbf{Toward value-aware evaluation.}
The empirical results support a shift from score-centric ranking to value-aware agent evaluation.
Stronger models raise the ceiling, but harness design determines how much of that capability becomes reliable, efficient, and auditable task completion.
Future evaluation should therefore measure not only task success, but also resource use, latency, timeout behavior, recovery quality, safety constraints, and trace auditability.
This observation motivates the value-aware objectives in Sec.~\ref{sec:future}, where success is evaluated jointly with cost, latency, risk, reliability, and process quality.

\section{Outlook and Future Directions}
\label{sec:future}

Future agent progress will require more than stronger foundation models or richer benchmarks.
We highlight three coupled directions: value-aware evaluation that accounts for success, cost, latency and safety; agent-native training that moves beyond planning and tool use toward verification and recovery; and harness design that adapts foundation models to task-specific tools, contexts, and constraints.
Together, they connect near-term harness engineering with longer-term efforts to internalize reliable interaction, adaptation, and self-improvement into agent models.

\subsection{From Score to Value-Aware Agent Optimization}
\label{sec:future:value-aware}

Current agent leaderboards~\cite{swebenchleaderboard,terminalbenchleaderboard} are largely score-centric: systems are ranked by task success, while API cost, latency, safety, and trace quality are secondary or missing.
This is useful for frontier comparison but incomplete for deployment, where cost-controlled, reliability-oriented, procedure-aware, and enterprise evaluation all point toward multi-dimensional agent quality~\cite{kapoor2024ai,liu2025costbench,gupta2026reliabilitybench,cao2026beyond,mehta2025beyond,opensquilla2026}.

A natural way to express this shift is to move from raw task success to value-aware agent optimization.
Let $\tau \sim \mathcal{D}$ denote a task instance, and let
$z=\mathrm{Run}(\tau;\mathcal{M},\mathcal{H},\omega)$ denote the execution trace produced by model $\mathcal{M}$ and harness $\mathcal{H}$ under stochasticity $\omega$.
For a trace $z$, let $S(z)\in\{0,1\}$ indicate task success, $C(z)$ execution cost, $L(z)$ latency, and $R(z)$ safety or compliance risk.
Cost may include token/API usage, tool calls, compute, or infrastructure; latency may be wall-clock time or interaction steps; risk may come from policy violations, safety checkers, or human audits.
Let $V(\tau)\geq0$ denote task utility, such as user value, scientific value, priority, or risk-adjusted importance.
The success probability of a model--harness pair can then be estimated from repeated runs as
\begin{equation}
P_{\mathrm{succ}}(\tau;\mathcal{M},\mathcal{H})
=
\mathbb{E}_{\omega}\!\left[S\!\left(\mathrm{Run}(\tau;\mathcal{M},\mathcal{H},\omega)\right)\right].
\end{equation}
Let $Q_{\mathrm{proc}}(z)\in[0,1]$ summarize process quality, including trace inspectability, verifier use, recovery behavior, provenance quality, and policy compliance.
Let $\mathrm{Rel}_k(\tau;\mathcal{M},\mathcal{H})$ denote repeated-run reliability estimated from $k$ runs, for example through consistency, pass@$k$, or stress-test reliability.
Instead of maximizing success alone, value-aware optimization can be written as
\begin{equation}
\begin{aligned}
\max_{\mathcal{M},\mathcal{H}} \quad
& \mathbb{E}_{\tau \sim \mathcal{D}}
\left[
V(\tau)\,
P_{\mathrm{succ}}(\tau;\mathcal{M},\mathcal{H})\,
\bar{Q}_{\mathrm{proc}}(\tau;\mathcal{M},\mathcal{H})
\right] \\
\mathrm{s.t.} \quad
& \mathbb{E}_{\tau,\omega}\!\left[C(z)\right] \leq B_C, \;
\mathrm{Quantile}_{p}\!\left(L(z)\right) \leq B_L, \\
& \mathbb{E}_{\tau,\omega}\!\left[R(z)\right] \leq \epsilon, \;
\mathbb{E}_{\tau}\!\left[\mathrm{Rel}_k(\tau;\mathcal{M},\mathcal{H})\right] \geq \rho .
\end{aligned}
\end{equation}
where $\bar{Q}_{\mathrm{proc}}(\tau;\mathcal{M},\mathcal{H})=\mathbb{E}_{\omega}[Q_{\mathrm{proc}}(z)]$.
This is not a universal leaderboard score; it makes the deployment target explicit by coupling task value with cost, latency, risk, and reliability constraints.
The trade-off is task-dependent: high-value or high-risk tasks may justify stronger verification, while routine high-frequency tasks favor cheaper models, shorter trajectories, and stricter stopping policies.

Let $\tilde{C}(z)=C(z)/B_C$, $\tilde{L}(z)=L(z)/B_L$, and $\tilde{R}(z)=R(z)/\epsilon$ be normalized cost, latency, and risk.
A complementary value-density objective is
\begin{equation}
\begin{aligned}
\mathrm{VD}_{\alpha,\beta,\gamma}
&=
\mathbb{E}_{\tau,\omega}
\left[
V(\tau)S(z)Q_{\mathrm{proc}}(z)D(z)^{-1}
\right],\\
D(z)
&=
(1+\tilde{C}(z))^{\alpha}
(1+\tilde{L}(z))^{\beta}
(1+\tilde{R}(z))^{\gamma}.
\end{aligned}
\end{equation}
where $\alpha$, $\beta$, and $\gamma$ control the penalties on cost, latency, and risk.
This is a family of deployment-specific utilities rather than a fixed metric.
Different deployments can instantiate it differently: high-value tasks may tolerate stronger verification, high-frequency workflows may penalize latency and cost, and safety-critical settings may replace soft risk penalties with hard constraints.
The same traces also support simpler reports, such as cost per effective success or latency per successful task, which distinguish systems with similar success rates but different runtime profiles.

From this perspective, harness engineering is a resource-allocation problem.
The harness chooses models, context, memory access, tools, retries, verifiers, stopping rules, and human escalation.
Model routing, context compression, cache reuse, verifier selection, recovery policy, and early stopping determine useful progress per unit cost, rather than being mere implementation details.
Future benchmarks should therefore report success together with token/API cost, tool calls, retries, 95th-percentile (P95) latency, recovery behavior, policy violations, and trace auditability.

\subsection{Learning to Verify, Recover, and Adapt}
\label{sec:future:learning}

The value-aware view also suggests a path for agent learning.
Execution traces are not only evaluation records; they contain outcomes, cost, tool calls, verifier signals, recovery attempts, policy violations, and feedback.
Future agents should learn not only to plan and act, but also to verify intermediate states, diagnose failures, recover from local errors, and adapt across tasks.

A useful abstraction is a constrained self-evolution loop.
Let $\theta_t$ denote the model parameters and $\phi_t$ denote the harness configuration at iteration $t$.
Running the agent on tasks from $\mathcal{D}$ produces traces $\mathcal{Z}_t$, from which the system extracts an evidence set $\mathcal{E}_t$ containing outcomes, failure modes, verifier results, cost profiles, and safety events.
An update operator $U$ may then change the model, the harness, or both:
\begin{equation}
(\theta_{t+1},\phi_{t+1})
=
\mathrm{VerifyRetain}\!\left(
U(\theta_t,\phi_t,\mathcal{E}_t)
\right).
\end{equation}
Here $\mathrm{VerifyRetain}$ keeps an update only if it passes held-out tasks, regression tests, process checks, and safety constraints; otherwise it is rejected or rolled back.
The expression is not a fixed algorithm; it makes the control structure explicit: reliable self-evolution must couple experience extraction, credit assignment, modification, and validation.

Existing agent-native training mostly advances the model side of this loop.
Interactive RL and environment-based training reduce train--test mismatch in web, computer-use, and software-engineering agents~\cite{qi2024webrl,lai2025computerrl,zeng2026davinci,yang2025kimi}.
Self-evolving systems further treat interaction experience as a reusable learning signal for self-questioning, attribution, online adaptation, and reward-free exploration~\cite{wu2025evolver,zhai2025agentevolver,karten2026continual,zhang2026training}.
Together, these works suggest that verification, recovery, and adaptation should become trainable behaviors, not only prompt-induced routines.

Parameter updates alone cannot absorb all runtime bottlenecks.
Many failures arise from harness choices: observation format, action granularity, memory retrieval, or verifier timing.
Agentic Harness Engineering (AHE) makes this harness-side path concrete by freezing the base model and evolving coding-agent harness components through observability-driven feedback~\cite{lin2026agenticHarnessEngineering}.
Its key lesson is that self-evolution must be observable and falsifiable: components should be explicit and revertible, traces should be distilled into evidence, and proposed changes should make predictions that later outcomes can check.

The long-term direction is co-evolution of models and harnesses.
Optimized harnesses produce better traces for training; trained models internalize recurring verification and recovery patterns; the resulting model changes which harness structure is optimal.
This introduces risks such as benchmark overfitting, incorrect failure attribution, stale memory, and unsafe runtime modification.
Agent-native training therefore does not eliminate the harness; it turns the harness into a training environment, evidence pipeline, verifier, and governance layer, with held-out evaluation, ablations, audit logs, rollback, and human approval for high-impact changes.

\begin{table*}[t]
\centering
\caption{\small{Representative harness designs behind agent benchmark performance. The table is not exhaustive; it highlights harness families that explain the empirical patterns in Sec.~\ref{sec:sec7:eval}.}}
\label{tab:harness-designs}
\fontsize{7pt}{7.2pt}\selectfont
\setlength{\tabcolsep}{1.6pt}
\renewcommand{\arraystretch}{0.78}
\resizebox{\textwidth}{!}{%
\begin{tabular}{@{}
>{\RaggedRight\arraybackslash}m{3.35cm}
>{\RaggedRight\arraybackslash}m{2.35cm}
>{\RaggedRight\arraybackslash}p{4.75cm}
>{\RaggedRight\arraybackslash}p{3.55cm}
>{\RaggedRight\arraybackslash}p{3.50cm}@{}}
\toprule
\textbf{Harness} & \textbf{Control style} & \textbf{Key design} & \textbf{Strength} & \textbf{Typical limitation} \\
\midrule
\rowcolor{gray!10}
SWE-agent~\cite{yang2024sweagent}
& ReAct-style loop
& LM interacts with shell/editor tools and iteratively inspects, edits, and tests code.
& Simple and general; strong baseline for real GitHub issues.
& Can be unstable and token-expensive on long debugging trajectories. \\

mini-SWE-agent~\cite{minisweagent2025}
& Minimal tool loop
& Roughly 100-line scaffold exposing compact shell/edit actions and leaving most orchestration to the model.
& High transparency; strong controlled comparisons across frontier backbones.
& Relies on the model to manage planning, context, and recovery. \\

\rowcolor{gray!10}
Agentless~\cite{xia2025demystifying}
& Fixed pipeline
& Staged localization, repair generation, and patch selection without a fully autonomous interaction loop.
& Stable, cheaper, and easier to reproduce.
& Less adaptive when the issue requires exploratory debugging. \\

AutoCodeRover~\cite{zhang2024autocoderover}
& Search-guided repair
& Repository-aware code search, AST-level localization, patch generation, and validation.
& Strong at locating relevant files/functions before editing.
& Depends heavily on localization quality and repo search signals. \\

\rowcolor{gray!10}
OpenHands + CodeAct~2.1~\cite{wang2024openhands}
& General runtime agent
& Full software-engineering runtime with shell, file editing, browser/tools, and iterative execution.
& Flexible for broad coding tasks and long interactions.
& Higher orchestration cost and larger action space. \\

PatchPilot~\cite{li2025patchpilot}
& Structured repair workflow
& Reproduction, localization, generation, validation, and refinement are organized as a controlled pipeline.
& Good cost-performance trade-off; validation-focused.
& Less open-ended than fully interactive agents. \\

\rowcolor{gray!10}
Codex / Claude Code~\cite{openai2026harness,claudecode2025}
& Managed coding agent
& Proprietary coding runtime tightly couples model, code execution, editing, and task management.
& High end-to-end coding performance with productized recovery and state management.
& System details are less transparent than open-source harnesses. \\

Meta-Harness~\cite{lee2026metaharness}
& Search-optimized harness
& Treats prompts, tools, and runtime policies as a searchable harness design space.
& Directly optimizes the harness rather than only the model.
& Adds search cost and can overfit to benchmark-specific feedback. \\

\rowcolor{gray!10}
SageAgent / OpenSage~\cite{li2026opensage,terminalbenchleaderboard}
& Generated agent scaffold
& Uses a self-programming agent-generation engine to produce and refine executable agent scaffolds.
& Strong Terminal-Bench results with relatively low observed runtime.
& Public leaderboard evidence is observational rather than a controlled ablation. \\

Terminus~2~\cite{terminalbenchleaderboard,terminalbenchsubmissions}
& Reference terminal harness
& Terminal-Bench native scaffold with shell execution, task state, verifier feedback, and benchmark constraints.
& Useful anchor for cross-model and cross-harness terminal comparisons.
& Can expose high timeout rates on difficult interactive tasks. \\

\rowcolor{gray!10}
Terminus-KIRA~\cite{krafton2026terminuskira,terminalbenchleaderboard}
& Terminal-native agent
& Tool-calling, terminal interaction, completion checks, and verification-oriented execution.
& Strong for terminal-bench style tasks requiring environment manipulation.
& Performance depends on robust task completion detection. \\

Mux~\cite{terminalbenchleaderboard,terminalbenchsubmissions}
& Lightweight terminal harness
& Minimal terminal-oriented scaffold for executing and verifying tasks.
& Simple and relatively transparent.
& Weaker planning and recovery compared with richer runtimes. \\

\rowcolor{gray!10}
TongAgents~\cite{terminalbenchleaderboard,terminalbenchsubmissions}
& Terminal agent system
& Submission-level terminal harness combining command execution, state tracking, and completion control.
& Strong observed Gemini 3.1 Pro Terminal-Bench result.
& Design details are less documented than paper-backed harnesses. \\
\bottomrule
\end{tabular}%
}
\end{table*}

\subsection{Harness Generalization Versus Specialization}
\label{sec:future:generalization}

The previous two directions raise a systems question: should a harness be reusable across tasks or specialized for one environment?
The answer depends on which harness layer is being considered.
Tracing, sandboxing, permission control, artifact storage, budget management, model routing, and basic tool protocols can form a reusable substrate.
Observation shaping, action abstraction, memory policy, verifier design, and recovery strategy are more often tied to the task pressure profile.

This distinction explains why realistic benchmarks are both specialized and compositional.
Software-engineering benchmarks stress verification and reversible execution~\cite{jimenez2024swe}; web and GUI benchmarks stress grounding, session state, and safe action selection~\cite{zhou2024webarena,xie2025osworld}; workplace and cross-service benchmarks stress coordination across heterogeneous tools and failure modes~\cite{xu2026theagentcompany,long2026liveclawbench}.
These settings place their primary bottlenecks on different harness layers, as discussed in Sec.~\ref{sec:tasks}.
A generic harness improves reuse and lowers engineering cost, but may provide weak inductive bias for the target bottleneck; a specialized harness can improve peak performance, but may reduce transferability and overfit to a benchmark-specific surface.

A practical direction is layered design: a general substrate plus domain-specific adapters.
The substrate provides logging, isolation, permission control, persistence, cost accounting, standardized tool access, and auditability.
Adapters define observations, actions, verifiers, memory policies, and retry, rollback, stopping, or escalation rules.
Protocols such as MCP and A2A reduce connector fragmentation and improve interoperability~\cite{mcp2025,google2025a2a}, but protocol standardization is not the same as harness generalization.
The harness must still decide what to expose, which actions to allow, how to verify outcomes and  recover from failure.

Future evaluations should test whether harness improvements transfer across task distributions, model families, and adapter choices, not only whether they raise one benchmark score.
Useful evidence includes same-model different-harness comparisons, component ablations, adapter replacement tests, held-out tasks, cross-domain transfer, and runtime profiles.
For self-evolving harnesses, this separates benchmark-specific tuning from reusable gains in tracing, memory compression, tool abstraction, verifier selection, or recovery policy~\cite{lin2026agenticHarnessEngineering}.
The long-term goal is modular and pressure-aware harness design: reusable substrates provide stability, observability, and governance, while domain adapters inject task-specific observation, action, verification, and recovery biases.
% \begin{table*}[t]
% \centering
% \caption{SWE-bench Verified performance across harnesses and primary models. Each model reports resolved rate and cost per instance.}
% \label{tab:swebench-harness-model-cost}
% \scriptsize
% \setlength{\tabcolsep}{3.5pt}
% \renewcommand{\arraystretch}{1.15}
% \begin{tabular}{@{}lcc cc cc@{}}
% \toprule
% \multirow{2}{*}{\textbf{Harness}}
% & \multicolumn{2}{c}{\textbf{GPT-4o}}
% & \multicolumn{2}{c}{\textbf{Claude 3.5 Sonnet}}
% & \multicolumn{2}{c}{\textbf{Claude 3.5 Sonnet (20241022)}} \\
% \cmidrule(lr){2-3}\cmidrule(lr){4-5}\cmidrule(l){6-7}
% & \textbf{Res.} & \textbf{Cost}
% & \textbf{Res.} & \textbf{Cost}
% & \textbf{Res.} & \textbf{Cost} \\
% \midrule
% SWE-agent        & 23.2 & --   & 33.6 & --   &      &      \\
% AutoCodeRover-v2 & 38.4 & --   &      &      &      &      \\
% Agentless        & 38.8 & --   &      &      & 50.8 & 1.19 \\
% Tools            &      &      &      &      & 49.0 & --   \\
% AutoCodeRover    &      &      & 51.8 & 4.50 &      &      \\
% OpenHands        &      &      & 53.0 & 0.78 &      &      \\
% PatchPilot       &      &      & 53.6 & 0.99 &      &      \\
% \bottomrule
% \end{tabular}
% \end{table*}

\section{Conclusion}
\label{sec:conclusion}

This survey argued that the development of LLM-based agents is best understood as an evolution across four paradigms: prompt engineering, agentic workflows, harness engineering, and agent-native training. The key systems insight is that agent performance is increasingly governed by the interaction between model and runtime rather than by model capability in isolation.
The harness perspective helps explain why similar base models can behave so differently once deployed in different environments. It also clarifies why recent progress has depended so heavily on context management, verification, tool design, orchestration, and recovery. At the same time, the rise of RL for agentic behavior suggests that some of this external scaffolding will gradually be internalized into model parameters.

The field is still early. Reliability remains below deployment needs in many realistic settings, evaluation is still only partially aligned with real use, and the boundary between model design and system design is still being renegotiated. But the direction is clear: agent engineering has moved beyond prompt craft and into the study of full systems. Understanding that shift is essential for both building better agents and evaluating their progress responsibly.

% \appendices
% \input{sections/app_literature_map}

\ifCLASSOPTIONcaptionsoff
  \newpage
\fi

\small{
\bibliographystyle{IEEEtran}
\bibliography{ref}
}

\end{document}